\documentclass[preprint,review,12pt,authoryear]{elsarticle}

\usepackage{amssymb}
\usepackage{amsmath}

\usepackage[left=1in, right=1in, top=1in, bottom=1in, textwidth=6.5in, textheight=9in]{geometry}

\usepackage{hyperref}
\usepackage{multirow}
\usepackage{booktabs}
\usepackage{algorithmic}
\usepackage{algorithm}
\usepackage{graphicx}

\usepackage{bbding}
\usepackage{pifont}
\usepackage{subcaption}
\usepackage{placeins}
\usepackage{lineno}
\usepackage{orcidlink}

\journal{ISPRS Journal of Photogrammetry and Remote Sensing}

\renewcommand{\cite}{\citep}
\begin{document}

\begin{frontmatter}



\title{Towards Lifelong Aerial Autonomy: Geometric Memory Management for Continual Visual Place Recognition in Dynamic Environments\tnoteref{funding}} 
\tnotetext[funding]{This work was supported in part by the Tsinghua-Toyota Joint Research Fund, in part by the State Key Laboratory of Autonomous Intelligent Unmanned Systems under Grant ZZKF2025ZD-2-2, in part by the Beijing Natural Science Foundation under Grant L252095, and in part by the National Natural Science Foundation of China under Grant 62403269.}


\cortext[cor1]{Corresponding authors.}

\author[thu]{Xingyu Shao} 
\ead{shao-xy21@mails.tsinghua.edu.cn}

\author[bistu]{Zhiqiang Yan}
\ead{2024020267@bistu.edu.cn}

\author[bistu]{Liangzheng Sun}
\ead{2023030031@bistu.edu.cn}

\author[thu]{Mengfan He}
\ead{hmf21@mails.tsinghua.edu.cn}

\author[thu]{Chao Chen}
\ead{chen-c@mail.tsinghua.edu.cn}

\author[bitauto]{Jinhui Zhang}
\ead{zhangjinh@bit.edu.cn}

\author[thu,bitaero]{Chunyu Li\corref{cor1}}
\ead{chunyuli@bit.edu.cn}

\author[thu]{Ziyang Meng\corref{cor1}}
\ead{ziyangmeng@tsinghua.edu.cn}

\affiliation[thu]{organization={Department of Precision Instrument, Tsinghua University},
            addressline={No. 30, Shuangqing Road, Haidian District}, 
            city={Beijing},
            postcode={100084}, 
            country={China}}

\affiliation[bistu]{organization={College of Instrument Science and Opto-electronics Engineering, Beijing Information Science and Technology University},
			addressline={No. 12, Xiaoying East Road, Haidian District}, 
			city={Beijing},
			postcode={100192}, 
			country={China}}

\affiliation[bitauto]{organization={Key Laboratory of Complex System Intelligent Control and Decision Making, Beijing Institute of Technology},
	addressline={No. 5, South Zhongguancun Street, Haidian District}, 
	city={Beijing},
	postcode={100081}, 
	country={China}}

\affiliation[bitaero]{organization={School of Aerospace Engineering, Beijing Institute of Technology},
            addressline={No. 5, South Zhongguancun Street, Haidian District}, 
            city={Beijing},
            postcode={100081}, 
            country={China}}

\begin{abstract}
Robust geo-localization under changing environmental and operational conditions is critical for long-term aerial autonomy. Aerial visual place recognition (VPR) commonly uses pre-acquired remote-sensing imagery of the intended operating area, so the geographic label space can remain fixed while successive airborne missions introduce substantial visual distribution shifts. Continual adaptation to these shifts can cause catastrophic forgetting. We therefore formulate aerial VPR as a mission-based domain-incremental learning (DIL) problem and develop a heterogeneous memory framework. Before sequential adaptation, the satellite reference dataset is used once to train the initial model and construct a static satellite exemplar memory; a bounded replay buffer then retains selected airborne observations across missions. For replay management, we compare loss- and diversity-based selection criteria and introduce DBS-Hybrid, which combines prototype-based diversity trimming with representative-first feature-space coverage. Experiments on 21 visible and infrared UAV missions evaluate generalization to held-out missions, immediate adaptation, and knowledge retention. Under the primary Forward mission order, DBS-Hybrid achieves the highest mean final average accuracy, generalization, and knowledge retention among the evaluated methods, improving over the Random baseline by $5.06$, $5.32$, and $6.33$ percentage points, respectively, and improving backward transfer from $-6.41\%$ to $1.07\%$. Across five additional random mission orders, DBS-Hybrid ranks second in mean final average accuracy, backward transfer, generalization, and knowledge retention. Overall, heterogeneous memory and diversity-aware replay provide an effective basis for continual aerial VPR in mapped operating areas.
\end{abstract}

\begin{graphicalabstract}
\centering
\includegraphics[width=\linewidth]{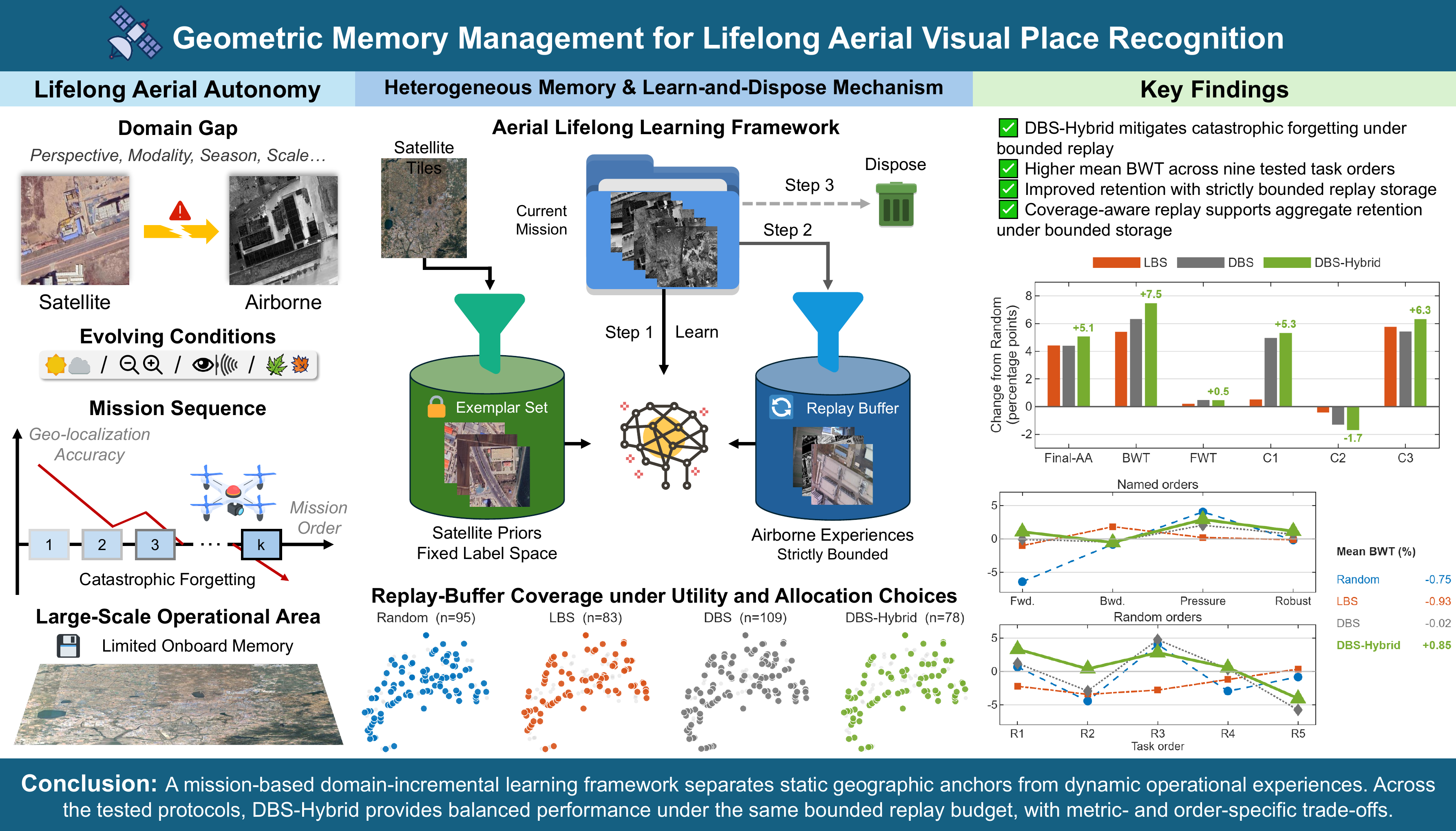}
\end{graphicalabstract}

\begin{highlights}
	\item A mission-based framework is proposed for aerial localization with fixed labels.
	\item Heterogeneous memory is proposed to separate satellite and airborne experience.
	\item Diversity-based replay retains representative and complementary observations.
	\item Three criteria are defined for generalization, adaptation, and knowledge retention.
	\item Diversity-aware replay improves retention over random replay under bounded memory.
\end{highlights}


\end{frontmatter}


\section{Introduction}
\label{sec:intro}

Robust localization is indispensable for autonomous unmanned aerial vehicles (UAVs) operating in global navigation satellite system (GNSS)-denied environments, as highlighted in recent comprehensive reviews~\cite{LATEEF2025402, NEX2022215}.
Visual place recognition (VPR), which determines location by matching onboard imagery against a georeferenced database, has emerged as a critical technology for achieving robust localization.
In particular, aerial robots must maintain localization over large geographic areas while handling changes in perspective, altitude, scene structure, and environmental conditions.

\begin{figure}[!t]
	\centering
	\includegraphics[width=0.8\linewidth]{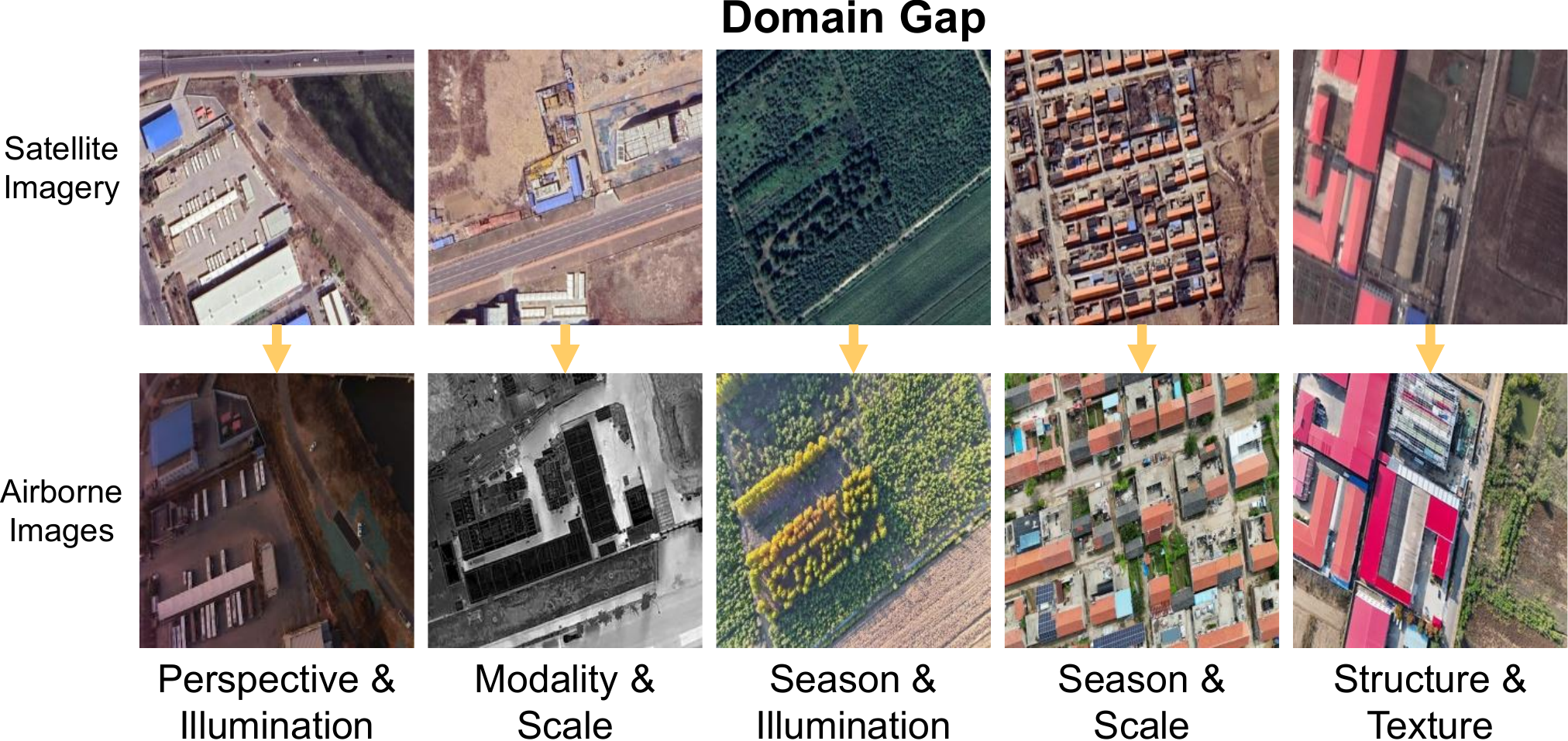}
	\caption{\textbf{Illustration of the substantial domain gaps in aerial VPR.} 
		The top row displays reference satellite imagery (satellite image tiles), while the bottom row shows the corresponding onboard aerial observations of the same geographic locations. 
		Each column highlights specific distributional shifts encountered during missions: (from left to right) varying \textit{perspective \& illumination}, distinct \textit{modality (VIS vs. IR) \& scale differences} (induced by flight altitude and camera parameters), complex \textit{seasonal} transitions and physical \textit{structure \& surface} evolution (e.g., structural modifications or land cover changes). 
		These distributional shifts motivate the mission-based domain-incremental learning (DIL) formulation studied in this work.}
	\label{fig:domain_gap}
\end{figure}

However, a model trained once may become mismatched to airborne observations that change across missions.
For aerial VPR, pre-acquired satellite imagery provides a geographic prior, while the airborne observation distribution evolves during long-term operation.
As illustrated in Figure~\ref{fig:domain_gap}, the visual appearance of the same geographic coordinate can change substantially due to seasonal transitions, altitude variations, illumination fluctuations, structural modifications, or sensor modality switches (e.g., from visible (VIS) to infrared (IR)). 
Traditional ``train-once, deploy-forever'' strategies cannot update their representations as these domain distributions change. 
Online fine-tuning enables adaptation to new mission data but may cause catastrophic forgetting, in which performance on previously learned missions degrades during subsequent adaptation. 
Continual learning (CL) provides a means to adapt sequentially while retaining knowledge acquired from previous missions.

While CL has been extensively explored in ground-based visual place recognition~\cite{airloop2022,bioslam2023,viper2025}, aerial VPR presents additional challenges.
First, limited onboard storage constrains the amount of mission experience that can be retained during large-area operation~\cite{moskalenko2025, geovins2025}.
A substantial domain gap also exists between airborne observations and satellite reference imagery.
As shown in Figure~\ref{fig:domain_gap}, this gap involves compound variations in viewpoint and altitude, illumination, sensor modality (VIS--IR), season, and scene structure~\cite{Khelifi2020DeepLearning, Shafique2022DeepLearningBased}.
These challenges are compounded by perceptual aliasing in repetitive aerial textures (e.g., urban grids), which may contain fewer distinctive landmarks than typical street-level observations~\cite{moskalenko2025}.
Second, standard VPR paradigms do not directly address sequential adaptation over repeated missions. Metric-learning approaches widely used in VPR can undergo embedding shifts, because adapting the feature extractor to a new domain changes the representations of previously learned locations~\cite{viper2025}. Retrieval-based systems may consequently require descriptor re-extraction for the stored reference database when the updated feature extractor is used for inference. Most robotic CL formulations introduce new locations over time~\cite{airloop2022,bioslam2023,viper2025}, corresponding to \textit{class-incremental learning (CIL)} when locations are treated as classes. Aerial missions within a mapped area present a different problem: satellite imagery defines the geographic labels in advance, while repeated missions change the visual appearance of those locations. This structure corresponds to \textit{domain-incremental learning (DIL)}~\cite{three_scenarios2019}, but has received little attention in long-term aerial VPR.

This paper formulates aerial VPR in a known and finite operating area as mission-based DIL.
A pre-acquired satellite map partitions the operating area into geographic grid cells and defines a fixed geographic label space before deployment. 
Successive airborne missions preserve the geographic classes while changing the input distribution through variations in viewpoint, illumination, season, scale, scene structure, and sensor modality.
The formulation therefore addresses domain-incremental rather than class-incremental learning.

We adopt the classification-based Divide \& Classify (D\&C) formulation~\cite{dc2023} because it provides a stable geographic label space during continual adaptation. The network predicts one of the predefined geographic grid cells, whose centre is converted to a geographic coordinate for geo-localization evaluation. Accordingly, the learning problem is formulated as geographic grid-cell classification rather than conventional gallery-ranking VPR. Generalization concerns airborne observations from held-out missions within the mapped operating area.

To support continual adaptation, we introduce a heterogeneous memory framework. Before the mission sequence, the complete satellite reference dataset is used once to train an initial model and construct a class-balanced static satellite exemplar memory. Each mission then starts from the preceding model and uses current-mission data together with the static satellite exemplar memory and a bounded replay buffer of historical airborne observations. After adaptation, only selected current-mission observations are retained in the replay buffer. This separation preserves the pre-acquired geographic prior while bounding the stored airborne experience.

Under the proposed ``Learn-and-Dispose'' pipeline, replay management determines which observations are retained in the bounded buffer. Loss-based selection (LBS) assigns high utility to samples with high training loss, whereas diversity-based selection (DBS) assigns high utility to samples with low feature-space redundancy. Min-Guar retains one high-utility representative for each geographic class and fills the remaining buffer slots by scalar utility. The LBS--DBS comparison therefore examines scalar utility under the same Min-Guar allocation. DBS-Hybrid instead retains one prototype-nearest representative per class, trims prototype-distant candidates within each class, applies greedy cosine $k$-center coverage, and backfills the budget when necessary. It follows the diversity route through prototype-based trimming and feature-space coverage.

The main contributions of this paper are summarized as follows:
\begin{itemize}
	\item We formulate aerial VPR in mapped operating areas as a mission-based DIL problem with a fixed geographic label space. The formulation distinguishes geographic grid-cell classification as the training objective from coordinate-based geo-localization evaluation and defines three complementary criteria for generalization, immediate adaptation, and knowledge retention.
	\item We develop a heterogeneous memory framework in which the satellite reference dataset establishes the initial model and a static satellite exemplar memory before sequential adaptation, while a bounded airborne replay buffer retains selected mission experience. For replay management, we introduce DBS-Hybrid, which combines prototype-based diversity trimming with representative-first greedy cosine $k$-center coverage and budget backfilling.
	\item We evaluate the framework on 21 VIS and IR UAV missions using multiple mission orders, random seeds, replay budgets, distance thresholds, and continual-learning metrics. Under the primary Forward order, DBS-Hybrid achieves the highest final average accuracy (FAA), Generalization (C1), Knowledge Retention (C3), and multi-threshold geo-localization recall among the evaluated methods; across five additional random mission orders, it ranks second in mean FAA, BWT, C1, and C3 and achieves the highest mean FWT.
\end{itemize}

The remainder of this article is organized as follows.
Section~\ref{sec:related_works} reviews related work in aerial VPR and continual learning.
Section~\ref{sec:problem_formulation} formally defines the mission-based DIL problem and describes the classification-based VPR architecture.
Section~\ref{sec:methodology} details the proposed heterogeneous memory framework, including the ``Learn-and-Dispose'' pipeline, replay selection and allocation, and the evaluation criteria.
Section~\ref{sec:experiments} introduces the mission-based aerial benchmark, describes the experimental implementation, and presents the quantitative results.
Section~\ref{sec:discussion} discusses the behavior and interpretation of the proposed selection strategies and analyzes their limitations.
Finally, Section~\ref{sec:conclusion} concludes the paper.

\section{Related Work}
\label{sec:related_works}


\subsection{From Retrieval to Classification in Aerial VPR}
\label{sec:rw_aerial_vpr}
Driven by advances in deep learning for remote sensing~\cite{MA2019166}, visual place recognition (VPR) commonly uses convolutional neural network (CNN)-based global descriptors such as NetVLAD~\cite{netvlad2018} and GeM~\cite{gem2019}. 
Recent aerial geo-localization systems also use Transformer backbones~\cite{denseuav2024,dinov22023}.
Recent foundation models such as DINOv2~\cite{dinov22023, salad2024} and DINOv3~\cite{simeoni2025dinov3} provide transferable visual representations that can support VPR across varying visual domains. 
Traditionally, aerial VPR is formulated as an image retrieval task, focusing on matching query images against a database via metric learning~\cite{denseuav2024, university16522020}. 
Early works established benchmarks such as PatternNet~\cite{ZHOU2018197} for content-based image retrieval in remote sensing, providing an important basis for subsequent retrieval research.
However, matching onboard imagery with satellite reference imagery presents a substantial challenge due to compound domain gaps, including variations in off-nadir angle and sensor modality~\cite{LI2025841}.

Beyond changes in network architecture, an alternative line of work formulates VPR as a large-scale classification task rather than an image retrieval problem.
Methods such as CosPlace~\cite{cosplace2022} and Divide \& Classify (D\&C)~\cite{dc2023} partition geographic space into discrete classes and train the model to predict a geographic class rather than directly optimizing embedding distances.
Prior studies have shown that this classification-based formulation can be effective in large-scale applications.
This discrete formulation also provides a fixed, stable geographic label space derived \textit{a priori} from a satellite map, which supports our formulation of continual aerial VPR as a DIL problem.

Public aerial VPR datasets such as VPAIR~\cite{vpair2022} and SUES-200~\cite{sues2002023} primarily evaluate static localization or retrieval under predefined protocols. They do not directly provide naturally ordered operational domains over a fixed geographic label space together with repeated historical-task evaluation required by the mission-based DIL formulation. Recent VPR methods such as AnyLoc~\cite{anyloc2024}, Revisit Anything~\cite{revisitanything2024}, SALAD~\cite{salad2024}, and dilated superpixel aggregation~\cite{zeng2025dsa} further demonstrate that representation and aggregation choices can substantially affect retrieval performance.

\subsection{Continual Learning: From CIL to DIL}
\label{sec:rw_cl}

Continual learning (CL) is a learning paradigm in which a model sequentially learns a series of tasks while retaining knowledge from previously learned tasks. 
Existing CL methods mainly mitigate catastrophic forgetting through three strategies: regularization~\cite{lwf2016, ewc2017}, parameter isolation~\cite{packnet2018}, and replay~\cite{er2019Chaudhry}. 
Replay-based methods interleave retained samples from previous tasks with data from the current task, providing a practical way to preserve historical knowledge during sequential adaptation. 
Notable examples include iCaRL~\cite{icarl2017}, which combines exemplar replay with distillation, and Dark Experience Replay++ (DER++)~\cite{derpp2020}, which enforces function-space consistency via logit matching.
More recently, unified domain incremental learning (UDIL) combines supervised replay, history-model distillation, and domain alignment through adaptive loss weights for memory-based DIL~\cite{NEURIPS2023_30d046e9}.

In the context of lifelong robotic localization, several frameworks have been proposed for changing environments.
For instance, AirLoop~\cite{airloop2022} targets lifelong loop closure detection, while BioSLAM~\cite{bioslam2023} introduces a bio-inspired dual-memory management system for visual SLAM. 
More recently, VIPeR~\cite{viper2025} introduced an adaptive mining strategy to select informative samples during model updates. 
These studies often consider robotic localization problems in which new places are progressively encountered over time, so the effective place set can expand during operation~\cite{openloris2020,mapillary2020}. 
When geographic places are represented as classes, progressively introducing new places corresponds to a class-incremental setting in which the label space expands over time~\cite{vandeVen2022three}.

However, many aerial operations (e.g., agricultural monitoring and powerline inspection) take place in known, finite geographic areas for which prior satellite maps are available.
For these operations, the challenge is not to discover new geographic classes but to adapt to changes in the visual appearance of the same mapped areas over time (e.g., from summer to winter or from VIS to IR).
This scenario corresponds to the \textit{domain-incremental learning (DIL)} paradigm~\cite{three_scenarios2019}, in which the label space remains fixed while the input distribution shifts.
Few studies have formulated long-term aerial VPR as a DIL problem.

\subsection{Memory Management and Sample Selection}
\label{sec:related_memory}

In replay-based continual learning, the choice of stored exemplars affects subsequent model updates under a limited memory budget~\cite{er2019Chaudhry}. 
Under its standard assumptions, reservoir sampling~\cite{randomsampling1985} provides a uniform sample of the observed stream, but it can retain redundant observations such as consecutive aerial frames~\cite{RainbowMemory2021}. 
How samples are selected is therefore an important design choice under a fixed memory budget.

\subsubsection{Optimization-based Selection}

One stream of research defines sample importance based on training dynamics. 
Methods such as Gradient-based Sample Selection (GSS)~\cite{GradientBased2019} and GCR~\cite{gcr2022} use gradient information to identify informative samples for replay.
Similarly, Toneva et al.~\cite{toneva2018an} characterized sample difficulty by tracking ``forgetting events'' during training and used these events to identify examples that are repeatedly forgotten.
These approaches are related to the principle of hard example mining~\cite{ohem2016, discriminativelearning2015}, which emphasizes samples that are difficult for the current model.
However, gradient-based methods can add computation on resource-constrained platforms. 
In this work, loss-based selection (LBS) uses the sample loss as a direct measure of sample difficulty without computing an additional gradient-based selection score.

\subsubsection{Geometric and Diversity-based Selection}

Another stream of research focuses on the geometric distribution of samples in the feature space. 
A representative approach is the Herding strategy, used in iCaRL~\cite{icarl2017} and subsequently adopted in continual-learning methods such as end-to-end incremental learning (EEIL)~\cite{eeil2018} and learning a unified classifier incrementally via rebalancing (LUCIR)~\cite{lucir2019}.
This strategy selects exemplars to approximate the class mean in feature space.

Mean-based selection naturally emphasizes central samples, but a single class mean may provide an incomplete description of multimodal or non-compact VPR feature distributions.
Geographic regions can exhibit heterogeneous feature distributions due to viewpoint variations (e.g., nadir vs. off-nadir), seasonal and illumination fluctuations, or physical structural changes~\cite{Sattler_2018_CVPR}; the arithmetic mean may then lie in a low-density region of the feature space~\cite{sener2018activelearning}.
Samples away from the class centre may therefore provide complementary feature-space information under domain shifts.

Related work in coreset selection~\cite{OnlineCoresetSelection2021} considers feature-space coverage as an alternative to purely mean-based representation.
In the context of mobile robotics, CoVIO~\cite{covio2023} applies a related principle to visual-inertial odometry (VIO) by pruning spatiotemporally redundant frames. 
Inspired by this principle, DBS uses pairwise feature similarity to reduce redundancy in the replay buffer.
The two proposed scores emphasize different sample properties: LBS ranks samples by loss, whereas DBS ranks them by feature-space redundancy.
Their controlled comparison examines how sample difficulty and feature-space redundancy affect replay selection under the same allocation strategy.

\section{Problem Formulation}
\label{sec:problem_formulation}

In this section, we formally define the aerial VPR problem considered in this work.
We first formulate this problem as mission-based domain-incremental learning (DIL) in Sec.~\ref{sec:m_problem_setup}. 
Subsequently, we describe the classification-based VPR architecture used throughout our framework in Sec.~\ref{sec:m_architecture}.

\subsection{Mission-Based DIL Problem Setup}
\label{sec:m_problem_setup}

Let $\Omega \subset \mathbb{R}^2$ denote the continuous geographic operating area (i.e., defined by UTM coordinate bounds). 
The visual input space is defined as $\mathcal{X}$, representing the set of all possible aerial and satellite images.
Unlike robotic localization problems in which new places are progressively introduced over time~\cite{airloop2022,bioslam2023,viper2025}, the aerial VPR problem considered here uses a fixed geographic label space.

Prior to deployment, the operating area $\Omega$ is discretized into $C$ distinct grid cells based on a pre-acquired satellite map $\mathcal{M}_{sat}$ covering the target region. 
This discretization establishes a fixed geographic label space $\mathcal{Y} = \{1, 2, \dots, C\}$, where each discrete label $y \in \mathcal{Y}$ corresponds to a specific geographic sub-region within $\Omega$.
From $\mathcal{M}_{sat}$, we acquire a satellite reference dataset $\mathcal{D}_{sat} = \{(x_i^{sat}, y_i)\}_{i=1}^{N_{sat}}$ composed of $N_{sat}$ near-nadir satellite image patches. 
These patches are generated by sliding-window cropping with spatial overlap and angular data augmentation to sample locations and orientations in the mapped area.
Here, the label $y_i \in \mathcal{Y}$ is assigned according to the grid cell containing the geographic center coordinate of the patch $x_i^{sat}$.
Before the mission sequence, $\mathcal{D}_{sat}$ is used once to train the initial model $f_{\Theta_0}$, where $\Theta_0$ denotes its learned parameters, and to construct the class-balanced static satellite exemplar memory $\mathcal{M}_{EX}\subset\mathcal{D}_{sat}$ described in Sec.~\ref{sec:m_dual_memory}.
During the subsequent mission sequence, the fixed $\mathcal{M}_{EX}$ provides the satellite-derived replay data.
The objective is to adapt a deep neural network's representations to mission-dependent shifts in the marginal input distribution $P_k(\mathcal{X})$, while the geographic label space $\mathcal{Y}$ remains fixed across missions.

Formally, we model the lifelong operation of the UAV as an ordered sequence of discrete missions $\mathcal{S} = (s_1, s_2, \dots, s_K)$, where $K$ denotes the total number of missions in the sequence.
At each sequential step $k \in \{1, \dots, K\}$, mission $s_k$ constitutes one discrete learning task, and the learner receives a mission-specific dataset $\mathcal{D}_k = \{(x_j, y_j)\}_{j=1}^{N_k}$, where $N_k$ denotes the total number of image-label pairs in $s_k$.
Here, each image $x_j$ is drawn from a mission-specific marginal input distribution $P_k(\mathcal{X})$ characterizing the visual appearance at step $k$, and $y_j \in \mathcal{Y}$ is its corresponding ground-truth geographic label.
Each mission dataset $\mathcal{D}_k$ is partitioned into a training set $\mathcal{D}_k^{train}$ and a within-mission test split $\mathcal{D}_k^{test}$ before learning. 
Under the evaluated protocol, raw training data from historical missions $\{s_1, \dots, s_{k-1}\}$ are not reused directly at step $k$. 
Historical mission information is retained through the preceding model parameters $\Theta_{k-1}$ and a limited-capacity experience replay (ER) buffer containing selected historical images and their geographic labels.

\subsection{Classification-Based VPR Architecture}
\label{sec:m_architecture}

Consistent with the fixed geographic label space of the DIL formulation, we adopt a classification-based VPR architecture following the Divide \& Classify (D\&C) paradigm~\cite{dc2023}.
The classifier predicts one of the predefined geographic cells, while satellite reference samples provide supervision tied to the fixed geographic map during continual adaptation.
For the mapped operating area considered here, the label space and classifier output dimension remain unchanged across missions.

\subsubsection{Grid Partitioning and Fixed Label Space}
The continuous operating area $\Omega$ is discretized into $C$ non-overlapping grid cells.
This process formulates VPR as a $C$-class classification problem. 
Crucially, since the grid layout is derived \textit{a priori} from the satellite map $\mathcal{M}_{sat}$, the label space $\mathcal{Y} = \{1, \dots, C\}$ remains fixed throughout the evaluated mission sequence.
Each class $y \in \mathcal{Y}$ is associated with a geographic centroid $\mathbf{p}_{y} = \operatorname{Center}(y)$, which is used to convert a categorical prediction into a geographic coordinate for geo-localization evaluation~\cite{dc2023}.

\subsubsection{Network Architecture}
\label{sec:network_arch}
We formulate the VPR model $f_\Theta$ as a composition of a deep feature extractor $\phi$ and a classifier head $\psi$, i.e., $f_\Theta(\cdot) = \psi(\phi(\cdot))$.
Given an input image $x \in \mathcal{X}$, the feature extractor $\phi: \mathcal{X} \to \mathbb{R}^D$ (typically comprising a backbone and an aggregation layer) maps the image to a global feature vector $\mathbf{z} = \phi(x) \in \mathbb{R}^D$, where $D$ denotes the embedding dimension.
Subsequently, the classifier head $\psi: \mathbb{R}^D \to \mathbb{R}^C$ projects this embedding onto the fixed label space, producing a probability vector $\mathbf{v}$ whose $y$-th component $v_y = p(y\mid x)$ gives the conditional probability for class $y \in \mathcal{Y}$.
The replay framework can be used with classifier heads that output scores over the fixed geographic label space; the experiments use the grouped classifier architecture specified in Sec.~\ref{sec:impl_details}.
This formulation is also used by the subsequent sample selection strategies: \textit{diversity-based selection (DBS)} operates on the feature representations produced by $\phi$, while \textit{loss-based selection (LBS)} uses the classification loss computed from the output of $\psi$.
The learning objective for any given batch is to minimize the cross-entropy (CE) loss between the predicted distribution and the ground-truth geographic grid label.

\section{Methodology}
\label{sec:methodology}

\begin{figure}[!t]
	\centering
	\includegraphics[width=0.7\linewidth]{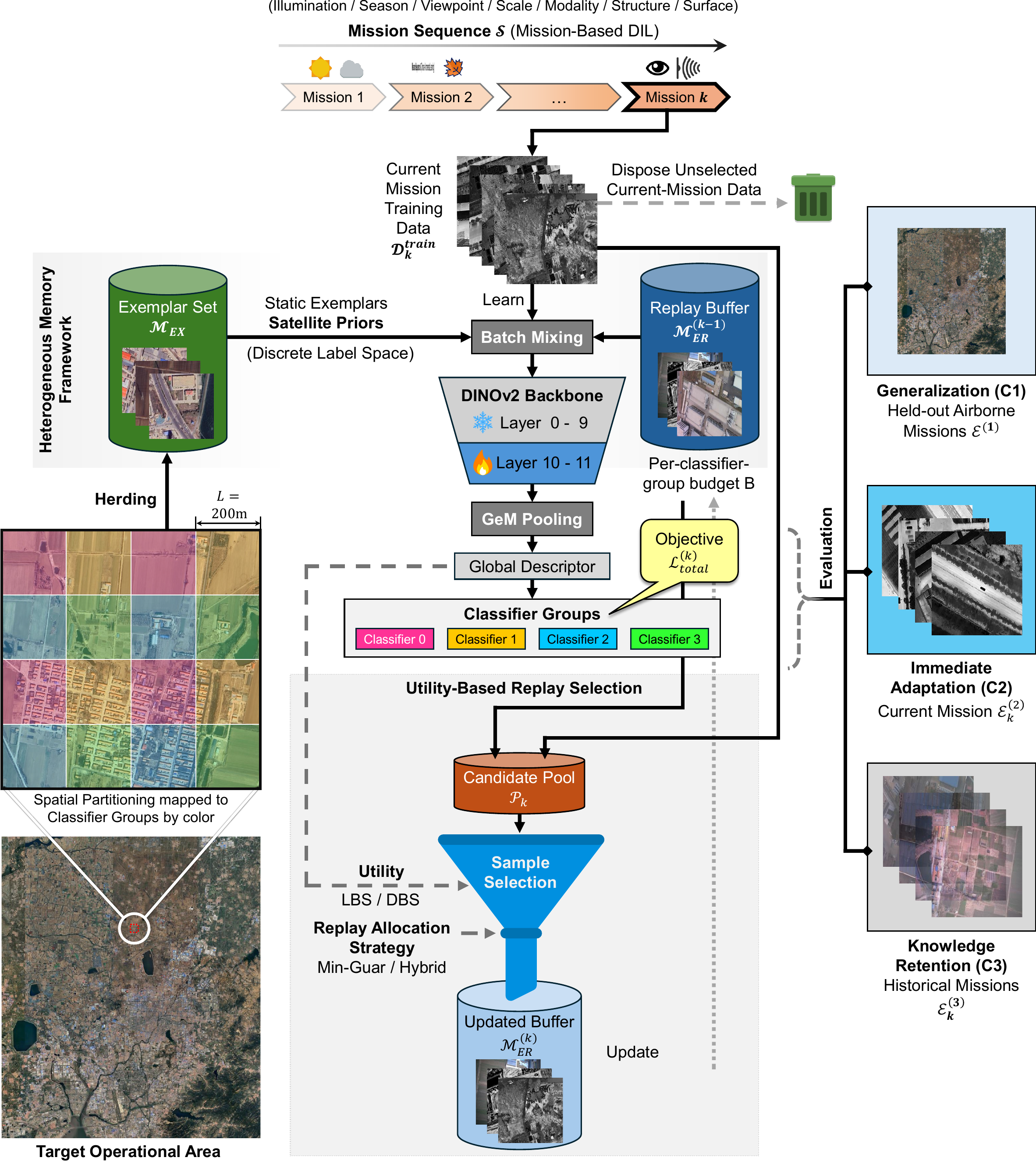}
	\caption{\textbf{Overview of the mission-based DIL framework.} \textbf{(Top)} At sequential step $k$, mission $s_k$ provides the current-mission training set $\mathcal{D}_k^{train}$ for adaptation; after learning, raw current-mission data are discarded except for samples selected into $\mathcal{M}_{ER}^{(k)}$. \textbf{(Left)} Before the mission sequence, the satellite reference dataset is used once to train the initial model $f_{\Theta_0}$ and construct the static satellite exemplar memory $\mathcal{M}_{EX}$. The heterogeneous memory framework then combines $\mathcal{M}_{EX}$ with the bounded experience replay buffer $\mathcal{M}_{ER}^{(k-1)}$. \textbf{(Center)} Mixed mini-batches pass through the DINOv2+GeM feature extractor and classifier groups, with the trainable components optimized using $\mathcal{L}_{total}^{(k)}$. \textbf{(Bottom)} The replay selection criterion and allocation strategy determine how the per-classifier-group budget $B$ is used. Min-Guar retains one high-utility representative per geographic class and ranks the residual candidates by scalar utility. DBS-Hybrid retains one prototype-nearest representative per class, trims the most prototype-distant class-wise candidates, applies greedy cosine $k$-center coverage, and backfills the budget when necessary. \textbf{(Right)} Evaluation considers generalization (C1) on held-out airborne missions ($\mathcal{E}^{(1)}$), immediate adaptation (C2) on the current mission ($\mathcal{E}_k^{(2)}$), and knowledge retention (C3) on historical missions ($\mathcal{E}_k^{(3)}$).}
	\label{fig:overall_framework}
\end{figure}

We study continuous adaptation under a bounded replay budget using a mission-based DIL framework.
As illustrated in Figure~\ref{fig:overall_framework}, the framework follows a ``Learn-and-Dispose'' pipeline that coordinates current-mission learning with the static satellite exemplar memory and the bounded ER buffer.
In this section, we first detail the operational pipeline (Sec.~\ref{sec:m_pipeline}) and the heterogeneous memory framework (Sec.~\ref{sec:m_dual_memory}), followed by the optimization objective (Sec.~\ref{sec:m_objective}). 
We then introduce the replay selection and allocation strategies (Sec.~\ref{sec:m_strategies}). 
Finally, we define the three evaluation criteria, namely generalization (C1), immediate adaptation (C2), and knowledge retention (C3) (Sec.~\ref{sec:m_criteria}).

\subsection{The ``Learn-and-Dispose'' Pipeline}
\label{sec:m_pipeline}

The operational process is divided into a sequence of discrete missions.
At mission step $k$, the learner uses the current-mission training set $\mathcal{D}_k^{train}$ together with the two memory components defined in Sec.~\ref{sec:m_dual_memory}.
The ``Learn-and-Dispose'' pipeline is then executed in three sequential phases:

\textit{First, the model adaptation phase updates the model parameters using mixed mini-batches of current-mission data, static satellite exemplars, and historical replay samples.}
Before discarding the raw current-mission data, the model updates its parameters from $\Theta_{k-1}$ to $\Theta_k$, with $\Theta_{k-1}=\Theta_0$ when $k=1$.
We construct mixed mini-batches from three sources: the current-mission training set $\mathcal{D}_k^{train}$, the static satellite exemplar memory $\mathcal{M}_{EX}$, and the previous ER buffer $\mathcal{M}_{ER}^{(k-1)}$ (detailed in Sec.~\ref{sec:m_dual_memory}).

\textit{Second, the buffer update phase applies a selection criterion and a class-wise allocation operator to construct the ER buffer for the next mission.}
For each classifier group $g$, we define the candidate pool as the union of its current-mission training samples and previous replay samples, $\mathcal{P}_{k,g}=\mathcal{D}_{k,g}^{train}\cup\mathcal{M}_{ER,g}^{(k-1)}$.
Let $q(x)$ denote the criterion-specific score assigned to candidate $x$; its definitions for the scalar-utility and Hybrid routes are given in Secs.~\ref{sec:m_scoring} and~\ref{sec:m_allocation}.
Let $B$ denote the maximum number of airborne replay samples stored for one classifier group. The updated group buffer $\mathcal{M}_{ER,g}^{(k)}$ is constructed by an allocation operator $\Phi$ under this per-classifier-group replay budget:
\begin{equation}
	\label{eq:buffer_update}
	\mathcal{M}_{ER,g}^{(k)} = \Phi \left( \mathcal{P}_{k,g}, q(\cdot), B \right)
\end{equation}
where $\Phi(\cdot)$ denotes the allocation operator (detailed in Sec.~\ref{sec:m_allocation}) that constructs the replay subset under the class-wise allocation rule and fixed budget $B$.

\textit{Finally, the disposal phase permanently removes the unselected raw current-mission data so that only the selected replay samples are retained for subsequent missions.}
Once the model parameters have been updated to $\Theta_k$ and the selected samples are stored in $\mathcal{M}_{ER}^{(k)}$, the remaining raw training data from mission $s_k$, i.e., $\mathcal{D}_k^{train} \setminus (\mathcal{D}_k^{train} \cap \mathcal{M}_{ER}^{(k)})$, are permanently removed.
This step bounds the ER buffer by the fixed capacity $B$ for each classifier group before mission $s_{k+1}$.
The static satellite exemplar memory $\mathcal{M}_{EX}$, satellite reference dataset $\mathcal{D}_{sat}$, and trained model parameters are separate storage components and are not included in this bounded ER-buffer claim.

\subsection{Heterogeneous Memory Framework}
\label{sec:m_dual_memory}

To mitigate catastrophic forgetting during mission-based continual adaptation, we develop a heterogeneous memory framework that separates a static satellite exemplar memory from a bounded experience replay buffer for historical airborne data.

\textit{The static satellite exemplar memory $\mathcal{M}_{EX}$ maintains an immutable, class-balanced subset of satellite reference images selected by herding~\cite{icarl2017}.}
The satellite reference dataset $\mathcal{D}_{sat}$ may contain substantially more images than are required during continual adaptation, so only a selected subset is retained in $\mathcal{M}_{EX}$.
Accordingly, we construct the static satellite exemplar memory $\mathcal{M}_{EX} \subset \mathcal{D}_{sat}$ prior to deployment.
Instead of random sampling, herding iteratively selects exemplars so that the mean of the selected feature representations approximates the empirical feature mean of each geographic class in $\mathcal{D}_{sat}$.
$\mathcal{M}_{EX}$ remains immutable throughout the evaluated mission sequence and provides a fixed satellite-derived representation prior during continual adaptation.

\textit{The experience replay buffer $\mathcal{M}_{ER}^{(k)}$ retains a bounded subset of airborne samples from historical missions to support knowledge retention during subsequent adaptation.}
Let $N$ denote the D\&C grouping modulus along each grid axis (defined in Sec.~\ref{sec:impl_details}); the model therefore has $N^2$ classifier groups. 
Let $\mathcal{M}_{ER,g}^{(k)}$ denote the replay buffer of group $g\in\{1,\ldots,N^2\}$ after mission $s_k$, and let $\mathcal{M}_{ER}^{(k)}=\bigcup_{g=1}^{N^2}\mathcal{M}_{ER,g}^{(k)}$.
Each group is bounded by $B$ samples, i.e., $|\mathcal{M}_{ER,g}^{(k)}|\le B$, so the complete ER buffer satisfies $|\mathcal{M}_{ER}^{(k)}|\le N^2B$.
During adaptation to mission $s_k$, the previous ER buffer $\mathcal{M}_{ER}^{(k-1)}$ provides historical replay samples.
The selection strategies in Sec.~\ref{sec:m_strategies} update each group buffer independently.

\subsection{Optimization Objective}
\label{sec:m_objective}

The learning objective combines cross-entropy (CE) losses computed from the current-mission training set and the two memory components.
For any labelled sample set $\mathcal{A}$, let $\overline{\mathcal{L}}_{CE}(\mathcal{A};\Theta_k)=\mathbb{E}_{(x,y)\sim\mathcal{A}}[\mathcal{L}_{CE}(f_{\Theta_k}(x),y)]$ denote its mean CE loss, where $f_{\Theta_k}(x)$ is the model prediction parameterized by $\Theta_k$ at step $k$, and $y\in\mathcal{Y}$ is the corresponding ground-truth geographic label.
The total loss $\mathcal{L}_{total}^{(k)}$ at mission step $k$ is the following weighted sum:
\begin{equation}
	\mathcal{L}_{total}^{(k)}=\overline{\mathcal{L}}_{CE}(\mathcal{D}_k^{train};\Theta_k)+\lambda_{EX}\overline{\mathcal{L}}_{CE}(\mathcal{M}_{EX};\Theta_k)+\lambda_{ER}\overline{\mathcal{L}}_{CE}(\mathcal{M}_{ER}^{(k-1)};\Theta_k).
\end{equation}
The first term adapts the model to the visual statistics of the current mission. 
The second term replays the static satellite exemplars in $\mathcal{M}_{EX}$ to retain satellite-derived supervision.
The third term replays selected samples from preceding missions in $\mathcal{M}_{ER}^{(k-1)}$ to reinforce previously learned decision boundaries.
Hyperparameters $\lambda_{EX}$ and $\lambda_{ER}$ control the contributions of the two replay sources relative to the current-mission term.

\subsection{Replay Selection and Allocation Strategies}
\label{sec:m_strategies}

A main challenge in the ``Learn-and-Dispose'' pipeline is to determine which samples to retain in the limited buffer $\mathcal{M}_{ER}^{(k)}$.
We address this problem in two steps: (1) defining the criterion used to select samples, and (2) applying an allocation operator $\Phi$ that constructs the retained subset under the replay budget.
For compact notation throughout this subsection, we omit the group index $g$ from the candidate pool and use $\mathcal{P}_k$ to denote a generic per-group pool $\mathcal{P}_{k,g}$ defined in Eq.~(\ref{eq:buffer_update}).

\subsubsection{Scoring Function Definition}
\label{sec:m_scoring}

The scoring function defines the utility assigned to each replay candidate. We consider two criteria: sample difficulty and feature-space diversity.

\textit{Loss-based selection (LBS) considers sample utility as learning difficulty, using training loss to identify and retain hard-to-recognize samples.}
At mission step $k$, let $x_i\in\mathcal{P}_k$ denote a candidate sample with ground-truth label $y_i\in\mathcal{Y}$. LBS defines its utility as
\begin{equation}
		\label{eq:lambda_LBS}
		\sigma_{LBS}(x_i) = \mathcal{L}_{CE}(f_{\Theta_k}(x_i), y_i).
\end{equation}
Here, $\mathcal{L}_{CE}(\cdot, \cdot)$ denotes the standard CE loss and $f_{\Theta_k}(x_i)$ denotes the model output after learning mission $s_k$.
High-loss samples can correspond to hard-to-recognize environmental conditions (e.g., low-light or occlusion). Prioritizing such ``hard examples'' follows the same general motivation as hard example mining, where difficult samples receive greater emphasis during learning.

\textit{Diversity-based selection (DBS) considers sample utility from a feature-space diversity perspective, favouring samples with low redundancy in the candidate pool and low similarity to their learned classifier prototypes.}
We define a redundancy score $R(x_i)$ that measures pairwise feature similarity within the candidate pool together with class-conditioned prototype similarity.
For every candidate $x_i\in\mathcal{P}_k$ with label $y_i$, the score comprises a \textit{global redundancy} term and a \textit{class-prototype} term:
\begin{equation}
	\label{eq:lambda_DBS}
	R(x_i) = \sum_{x_j \in \mathcal{P}_k \setminus \{x_i\}} \operatorname{Sim}(\phi(x_i), \phi(x_j)) + \lambda \cdot \operatorname{Sim}(\phi(x_i), \mathbf{w}_{y_i})
\end{equation}
where $\phi(\cdot)$ denotes the feature extraction mapping defined in Sec.~\ref{sec:network_arch}, $\operatorname{Sim}(\cdot,\cdot)$ denotes cosine similarity, $\mathbf{w}_{y_i}$ is the learned classifier row associated with class $y_i$, which serves as a discriminatively learned angular prototype, and $\lambda$ controls the contribution of the class-prototype term.
The first term measures feature-space redundancy relative to the candidate pool, while the second provides a class-conditioned preference based on proximity to the learned classifier prototype.
To select the most distinctive samples, we define the utility score as the negative redundancy:
\begin{equation}
		\label{eq:sigma_DBS}
		\sigma_{DBS}(x_i) = - R(x_i)
\end{equation}
Maximizing $\sigma_{DBS}$ (i.e., minimizing $R$) favours samples that are less redundant with the candidate pool and less similar to the corresponding classifier prototype. We use this score as an operational proxy for feature-space coverage in the replay candidate pool.

Before evaluating Eq.~(\ref{eq:lambda_DBS}), each feature $\mathbf{z}_i=\phi(x_i)$ and classifier row $\mathbf{w}_{y_i}$ is L2-normalized as $\widetilde{\mathbf{z}}_i=\mathbf{z}_i/\max(\|\mathbf{z}_i\|_2,\epsilon)$ and $\widetilde{\mathbf{w}}_{y_i}=\mathbf{w}_{y_i}/\max(\|\mathbf{w}_{y_i}\|_2,\epsilon)$, where $\epsilon>0$ prevents division by zero. Both terms therefore use dimensionless cosine similarities in $[-1,1]$.
For a candidate pool of size $n$, the pairwise term sums $n-1$ similarities and lies in $[-(n-1),n-1]$. The weighted class-prototype term lies in $[-\lambda,\lambda]$ for $\lambda\ge0$. The two terms are not normalized to equal numerical scales. With the experimental setting $\lambda=1$, the pairwise-redundancy term generally has the larger numerical scale as $n$ grows, while the prototype term provides a secondary class-conditioned preference.
For $n$ candidates with feature dimension $d$, dense DBS requires $O(n^2d)$ time and $O(n^2)$ pairwise-similarity storage. The measured per-group candidate pools have a median size of 202, a 95th percentile of 243.4, and a maximum of 247; selector runtime is reported in Sec.~\ref{sec:efficiency}.
Classification training makes each classifier row a class-conditioned decision direction in the normalized feature space, so DBS uses it as a learned angular prototype rather than an arithmetic class mean. Across 512 sampled classes, the median row-to-centroid cosine similarity is approximately 0.66. The per-class row-to-centroid cosine values obtained from the DBS and DBS-Hybrid models are strongly correlated (Pearson $r=0.9655$). Replacing the same 512 classifier rows with centroids estimated from at most 12 images reduces Recall@$300\,\mathrm{m}$ by 8.56 percentage points for DBS and 8.40 points for DBS-Hybrid. These measurements characterize the classifier row as a learned class direction distinct from the sampled arithmetic centroid.

\subsubsection{Replay Allocation Strategy}
\label{sec:m_allocation}

The allocation operator $\Phi$ constructs the updated group buffer $\mathcal{M}_{ER,g}^{(k)}$ from the candidate pool $\mathcal{P}_k$ subject to the replay budget $B$. The selection criterion determines which sample property is favoured, whereas the allocation structure determines how representatives, trimming, coverage, and backfilling distribute the available budget.

A naive approach is \textit{Global} allocation, which retains the highest-utility samples from the entire candidate pool without explicitly considering their geographic classes. Although this directly prioritizes $\sigma(x)$, it does not guarantee representation of the geographic classes represented in $\mathcal{P}_k$. Alternatively, \textit{Round-Robin} allocation~\cite{kleinrock1964} distributes the replay capacity as evenly as possible across the represented geographic classes without considering differences in their utility-score distributions.

To balance geographic class representation and sample utility, we define a geographic-class-constrained allocation operator termed \textit{Minimum Guarantee (Min-Guar)}. Let $\mathcal{Y}_{\mathcal{P}_k}$ denote the set of unique geographic grid labels represented in $\mathcal{P}_k$. Min-Guar first identifies the highest-utility representative of each represented class to form
\begin{equation}
	\mathcal{P}_{rep} =
	\left\{
	\underset{x \in \mathcal{P}_k,\,\operatorname{label}(x)=c}{\operatorname{argmax}}
	\sigma(x)
	\;\middle|\;
	c \in \mathcal{Y}_{\mathcal{P}_k}
	\right\},
	\label{eq:min_guar_rep}
\end{equation}
where $\operatorname{label}(x)\in\mathcal{Y}$ denotes the ground-truth geographic grid label of sample $x$. Let $\operatorname{TopK}(\mathcal{S},n)$ select up to $n$ samples with the highest utility scores from a set $\mathcal{S}$. Min-Guar instantiates the allocation operator in Eq.~(\ref{eq:buffer_update}) as
\begin{equation}
	\Phi_{\mathrm{MG}}(\mathcal{P}_k,\sigma,B)=
	\begin{cases}
		\mathcal{P}_{rep}\cup\operatorname{TopK}(\mathcal{P}_{res},B_{res}), & B\ge|\mathcal{Y}_{\mathcal{P}_k}|,\\
		\operatorname{TopK}(\mathcal{P}_{rep},B), & B<|\mathcal{Y}_{\mathcal{P}_k}|,
	\end{cases}
	\label{eq:min_guar}
\end{equation}
where $\mathcal{P}_{res}=\mathcal{P}_k\setminus\mathcal{P}_{rep}$ is the residual pool and $B_{res}=B-|\mathcal{Y}_{\mathcal{P}_k}|$ is the surplus capacity. When the replay budget accommodates all represented classes, Min-Guar first retains one representative per class and then fills the remaining capacity with the highest-utility samples from the residual pool. When the number of represented classes exceeds $B$, it retains the $B$ highest-utility samples from $\mathcal{P}_{rep}$.

The LBS and DBS strategies use the same Min-Guar allocation and differ only in their utility definitions: LBS combines $\sigma_{LBS}$ with $\Phi_{\mathrm{MG}}$, whereas DBS combines $\sigma_{DBS}$ with $\Phi_{\mathrm{MG}}$. Global and Round-Robin provide alternative allocation baselines for evaluating the effect of the geographic-class constraint.

Min-Guar guarantees class-level representation whenever the replay budget permits, but its surplus allocation ranks the remaining candidates only by the scalar utility $\sigma(x)$. In DBS-Hybrid, the Hybrid allocation retains the representative-first structure while adding feature-space coverage to the allocation of the remaining capacity. For each represented class $c$, DBS-Hybrid selects the candidate with the highest cosine similarity to the learned classifier prototype $\mathbf{w}_c$. We denote the resulting representative set by $\mathcal{P}_{rep}^{H}$, where the superscript $H$ identifies the Hybrid allocation:
\begin{equation}
	\mathcal{P}_{rep}^{H}=
	\left\{
	\underset{x\in\mathcal{P}_k,\,\operatorname{label}(x)=c}{\operatorname{argmax}}
	\operatorname{Sim}\!\left(\phi(x),\mathbf{w}_c\right)
	\;\middle|\;
	c\in\mathcal{Y}_{\mathcal{P}_k}
	\right\}.
	\label{eq:hybrid_rep}
\end{equation}
If $|\mathcal{P}_{rep}^{H}|>B$, DBS-Hybrid retains the representatives associated with the first $B$ class IDs and terminates. Otherwise, it retains the complete set $\mathcal{P}_{rep}^{H}$ and allocates the remaining capacity as follows.

Let $\mathcal{I}_c=\{i:\operatorname{label}(x_i)=c\}$ denote the indices of all candidates from represented class $c$. For DBS-Hybrid, the criterion-specific trimming score is
\begin{equation}
	q_H(x_i)=\operatorname{Sim}\!\left(\phi(x_i),\mathbf{w}_{y_i}\right),
	\label{eq:hybrid_score}
\end{equation}
where $q_H(x_i)$ is used only to determine class-wise trimming eligibility. DBS-Hybrid therefore uses prototype similarity rather than the complete scalar DBS utility in Eq.~(\ref{eq:sigma_DBS}) and removes the lowest-scoring class-wise tail from the residual candidates according to a trimming ratio $\rho$, yielding the trimmed candidate set
\begin{equation}
	\mathcal{P}_{\rho}^{H}=
	\bigcup_{c\in\mathcal{Y}_{\mathcal{P}_k}}
	\left\{
	x_i\in\mathcal{P}_k\setminus\mathcal{P}_{rep}^{H}:
	i\in\mathcal{I}_c,\;
		q_H(x_i)\ge Q_{\rho}\!\left(\{q_H(x_j)\mid j\in\mathcal{I}_c\}\right)
	\right\},
	\label{eq:hybrid_trim}
\end{equation}
where $Q_{\rho}$ denotes the empirical $\rho$-quantile computed from the trimming scores of all candidates in class $c$, including its representative. The threshold is computed from the complete class-wise candidate set, but trimming is applied only to $\mathcal{P}_k\setminus\mathcal{P}_{rep}^{H}$; representatives are therefore always retained. For DBS-Hybrid, this removes the largest prototype-distance tail because higher prototype similarity gives a higher $q_H$. The implementation value of $\rho$ and the handling of classes with very few candidates are specified in Sec.~\ref{sec:impl_details}.

Let $\mathcal{R}$ denote the retained set under construction. Starting from $\mathcal{R}=\mathcal{P}_{rep}^{H}$, DBS-Hybrid assigns the remaining replay slots by greedy cosine $k$-center selection over the normalized feature representations defined in Sec.~\ref{sec:m_scoring}:
\begin{equation}
	i^*=
	\underset{x_i\in\mathcal{P}_{\rho}^{H}\setminus\mathcal{R}}{\operatorname{argmax}}
	\min_{x_j\in\mathcal{R}}
	\left(1-\widetilde{\mathbf{z}}_i^{\mathsf T}\widetilde{\mathbf{z}}_j\right),
	\qquad
	\mathcal{R}\leftarrow\mathcal{R}\cup\{x_{i^*}\}.
	\label{eq:hybrid_kcenter}
\end{equation}
This update is repeated until the replay budget is reached or no trimmed candidate remains. If trimming leaves too few candidates, the excluded residual candidates are reintroduced by the deterministic backfill rule. The final set satisfies $|\mathcal{R}|=\min(B,|\mathcal{P}_k|)$ and defines $\Phi_{H}(\mathcal{P}_k,q_H,B)=\mathcal{R}$, where $\Phi_H$ denotes the Hybrid allocation operator. Thus, Hybrid combines prototype-nearest class representation, criterion-specific trimming, greedy feature-space coverage, and budget backfilling.

\subsection{Evaluation Criteria}
\label{sec:m_criteria}

We define three evaluation criteria for the mission-based DIL problem considered in this work.
Each mission dataset $\mathcal{D}_k$ is partitioned into $\mathcal{D}_k^{train}$ and $\mathcal{D}_k^{test}$ as defined in Sec.~\ref{sec:m_problem_setup}.
The criteria evaluate generalization, immediate adaptation, and knowledge retention using one fixed held-out set $\mathcal{E}^{(1)}$ and two mission-step-specific sets $\mathcal{E}_k^{(2)}$ and $\mathcal{E}_k^{(3)}$.

\textit{To map the classification output to a coordinate-based geo-localization result, we define a distance-bounded geo-localization accuracy metric.}
As defined in Sec.~\ref{sec:m_architecture}, let $f_\Theta: \mathcal{X} \to \mathbb{R}^C$ denote the neural network parameterized by $\Theta$.
At step $k$, the predicted geographic grid label for image $x$ is $\hat{y}_k(x) = \operatorname{argmax}_{y \in \mathcal{Y}} v_y$, where $v_y$ is the $y$-th element of the output vector $\mathbf{v} = f_{\Theta_k}(x)$.
Using the mapping $\operatorname{Center}: \mathcal{Y} \to \Omega$ defined in Sec.~\ref{sec:m_architecture}, the predicted label $\hat{y}_k(x)$ is converted to the geographic centroid coordinate of the corresponding grid cell.
Let $\mathcal{E}\subseteq\mathcal{X}$ be a non-empty set of airborne evaluation images. Each image $x\in\mathcal{E}$ is associated with a ground-truth coordinate $\mathbf{p}_{gt}(x) \in \Omega$.
A geo-localization prediction is considered successful if the Euclidean distance between the predicted centroid and the ground-truth coordinate does not exceed a distance threshold $\tau$.
Let $\mathbb{1}(\cdot)$ be an indicator function that returns $1$ if the distance condition holds and $0$ otherwise.
The distance-bounded geo-localization accuracy on $\mathcal{E}$ after learning mission $s_k$ is defined as
\begin{equation}
	\begin{split}
	Acc_k(\mathcal{E};\tau) =
	\frac{1}{|\mathcal{E}|}
	\sum_{x \in \mathcal{E}}
	\mathbb{1}\left(
	\left\|
	\operatorname{Center}\!\left(\hat{y}_k(x)\right)
	-
	\mathbf{p}_{gt}(x)
	\right\|_2
	\le \tau
	\right).
	\label{eq:distance_recall}
	\end{split}
\end{equation}
When $\tau$ is fixed by the evaluation protocol, we abbreviate $Acc_k(\mathcal{E};\tau)$ as $Acc_k(\mathcal{E})$. Based on this coordinate-based geo-localization metric, we define the following three evaluation criteria.

\textit{Criterion 1 (C1: Generalization) evaluates geo-localization on held-out airborne missions that do not participate in continual training.}
We define the fixed C1 evaluation set as
\[
\mathcal{E}^{(1)} = \mathcal{U},
\]
where $\mathcal{U}$ denotes the held-out mission set.
In the experiments, $\mathcal{U}$ contains 11 held-out airborne missions. 
Their airborne images are excluded from continual training, while the corresponding geographic labels and satellite exemplars are defined by the pre-acquired map.
C1 therefore measures transfer to held-out airborne appearances over the fixed geographic label space of the mapped operating area.

\textit{Criterion 2 (C2: Immediate Adaptation) evaluates the model's plasticity on the current mission using $\mathcal{E}^{(2)}_k$.}
Accordingly,
\[
\mathcal{E}^{(2)}_k = \mathcal{D}_k^{test}.
\]
C2 therefore reflects the model's immediate adaptation to the current mission at step $k$.

\textit{Criterion 3 (C3: Knowledge Retention) evaluates the model's stability against catastrophic forgetting on historical missions using $\mathcal{E}^{(3)}_k$.}
For $k > 1$, the C3 evaluation set comprises the union of test splits from all preceding missions:
\begin{equation}
	\mathcal{E}^{(3)}_k = \bigcup_{i=1}^{k-1} \mathcal{D}_i^{test}.
\end{equation}
C3 therefore measures knowledge retention on historical missions and excludes the current mission $\mathcal{D}_k^{test}$ from its evaluation set.

\textit{The three criteria complement standard continual-learning metrics by separately characterizing generalization, plasticity, and stability.}
Standard metrics such as forward transfer (FWT) and backward transfer (BWT) summarize performance changes across the mission sequence~\cite{gem2017}, whereas C1, C2, and C3 report held-out transfer, current-mission adaptation, and historical-mission retention on explicitly defined evaluation sets.

\section{Experiments}
\label{sec:experiments}

\subsection{Experimental Setup}
\label{sec:experimental_setup}

\subsubsection{The Heterogeneous Aerial Mission Benchmark}
\label{sec:dataset_desc}

\begin{table}[!t]
	\centering
	\caption{\textbf{Statistics of aerial missions used in this work.} Missions marked as ``CL'' form the mission sequence $\mathcal{S}$, while ``Held-out'' missions constitute the held-out mission set $\mathcal{U}$ used to evaluate generalization (C1).}
	\label{tab:dataset_statistics}
	\begin{tabular}{l c c c c}
		\toprule
		\textbf{Mission ID} & \textbf{Modality} & \textbf{\#Images} & \textbf{District} & \textbf{Split Usage} \\
		\midrule
		\texttt{JHT-01} & VIS & 63  & \textit{Chengyang} & CL \\
		\texttt{JHT-02} & VIS & 348 & \textit{Chengyang} & CL \\
		\texttt{JHT-04} & VIS & 257 & \textit{Chengyang} & CL \\
		\texttt{JHT-05} & IR & 257 & \textit{Chengyang} & CL \\
		\texttt{LSZ-01} & VIS & 275 & \textit{Jimo} & CL \\
		\texttt{LSZ-02} & VIS & 234 & \textit{Jimo} & Held-out \\
		\texttt{LSZ-03} & IR & 232 & \textit{Jimo} & Held-out \\
		\texttt{LSZ-04} & VIS & 215 & \textit{Jimo} & Held-out \\
		\texttt{LSZ-06} & VIS & 281 & \textit{Jimo} & CL \\
		\texttt{LSZ-07} & VIS & 284 & \textit{Jimo} & CL \\
		\texttt{LSZ-08} & IR & 214 & \textit{Jimo} & Held-out \\
		\texttt{TD-01-seq} & VIS & 193 & \textit{Jimo} & Held-out \\
		\texttt{TD-02-seq} & VIS & 237 & \textit{Jimo} & CL \\
		\texttt{TD-03-seq} & VIS & 347 & \textit{Jimo} & Held-out \\
		\texttt{TD-04-seq} & VIS & 328 & \textit{Jimo} & Held-out \\
		\texttt{TD-05-seq} & VIS & 468 & \textit{Jimo} & Held-out \\
		\texttt{TD-06-seq} & VIS & 167 & \textit{Jimo} & Held-out \\
		\texttt{TD-07-seq} & VIS & 189 & \textit{Jimo} & CL \\
		\texttt{TD-08-seq} & VIS & 194 & \textit{Jimo} & Held-out \\
		\texttt{TD-12-seq} & IR & 156 & \textit{Jimo} & Held-out \\
		\texttt{TD-13-seq} & IR & 500 & \textit{Jimo} & CL \\
		\bottomrule
	\end{tabular}
\end{table}

The mission-based aerial benchmark contains 21 UAV missions collected in the Chengyang and Jimo districts of Qingdao, China (Table~\ref{tab:dataset_statistics}).
Ten missions form the mission sequence $\mathcal{S}$, and the remaining 11 missions form the held-out mission set $\mathcal{U}$ used to evaluate generalization (C1).
The latter contains 2,748 queries, including 2,146 VIS and 602 IR images.

During acquisition, onboard global navigation satellite system (GNSS) coordinates were recorded in the video subtitles and associated with the extracted frames by timestamp. 
Coordinates use World Geodetic System 1984 / Universal Transverse Mercator (WGS 84 / UTM) zone 50N (EPSG:32650), with easting and northing in metres. 
The UAV frames were not geometrically rectified, orthorectified, or registered to the satellite imagery. 
Satellite tiles and UAV observations were associated through their projected coordinates. 
The geographic grid cell containing the UAV coordinate determines the ground-truth geographic label, while the centre of the predicted grid cell provides the coordinate-based geo-localization output.
The onboard GNSS coordinates provide the ground-truth positions; no separately surveyed image-registration ground truth is used.

Before temporal splitting, the ten sequential missions contain 2,691 UAV images. 
Within each mission, frames are sorted by acquisition order and divided into contiguous training and retention blocks at one boundary. Five adjacent frames on each side of the boundary are excluded, giving 100 excluded frames across the ten missions. 
The resulting split contains 1,294 training images and 1,297 retention queries. 
This block-wise procedure prevents temporally adjacent frames from being divided between learning and evaluation.
The 11 held-out missions remain disjoint from the 10-mission sequence $\mathcal{S}$ at the mission level.

In addition to environmental variations such as illumination and seasonal fluctuations and structural changes, the benchmark includes two additional perception challenges relevant to heterogeneous aerial platforms.
First, it features cross-modality shifts, including both visible spectrum (VIS) and infrared (IR) missions (e.g., \texttt{JHT-05}, \texttt{LSZ-03}). 
Second, the benchmark introduces a scale mismatch challenge.
Unlike the satellite reference imagery, which is cropped at a standardized fixed scale, the UAV imagery is collected at varying flight altitudes and with different sensor specifications (e.g., focal lengths and aspect ratios).
This produces different ground sample distances (GSDs) between the airborne observations and the satellite reference imagery.
The model must therefore accommodate scale variation between airborne observations and the fixed-scale satellite reference imagery.

\subsubsection{Mission Orders}
\label{sec:curriculum}

Mission order can affect continual-learning performance, a phenomenon known as the \textit{curriculum effect}~\cite{bell2022task}. 
Unless otherwise specified, experiments in Sec.~\ref{sec:main_results} use the Forward mission order.
To examine sensitivity to mission order, we evaluate four named orders of the mission sequence $\mathcal{S}$ as follows.
The \textit{Forward} order serves as the default mission order. It begins with VIS missions and ends with the 500-image IR mission \texttt{TD-13-seq}.
The \textit{Backward} order reverses the Forward order, placing the final IR mission first and the earlier VIS missions later in the sequence.
The \textit{Pressure} order places the 500-image IR mission \texttt{TD-13-seq} at the beginning of the sequence, introducing an early modality shift relative to the subsequent VIS-dominated missions.
The \textit{Robust} order uses one fixed shuffled arrangement of the ten missions to provide a non-chronological mission order.
The exact mission sequence for each order is provided in \ref{sec:appendix_curriculum}.

\subsubsection{Implementation Details}
\label{sec:impl_details}

All experiments are conducted using the PyTorch framework on a workstation equipped with a single NVIDIA RTX 4090 GPU. 
For the network architecture, we use DINOv2 ViT-B/14 as the backbone. 
To limit feature-space drift while allowing adaptation to evolving visual distributions, we freeze the first 10 Transformer blocks and fine-tune only the last two blocks. 
The extracted patch tokens are subsequently aggregated using a fixed, parameter-free generalized mean (GeM) pooling layer. 
Input images are resized to a canonical resolution of $224 \times 224$ pixels. 
During training, we apply standard data augmentations, including random resized cropping with a scale range of $[0.66,1.0]$ and normalization.

Following the D\&C~\cite{dc2023} paradigm, we discretize the continuous operating area $\Omega$ into square geographic grid cells. 
Since UAV observations can cover a comparatively large ground footprint, we set the grid-cell side length to $L=200\,\mathrm{m}$ to align the geographic label granularity with the ground coverage of aerial observations. 
This coarse-grained discretization has two advantages: 
(1) it reduces label ambiguity when a single aerial observation spans a large ground area; and 
(2) it constrains the cardinality of the geographic label space and consequently limits the size of the class-wise static satellite exemplar memory $\mathcal{M}_{EX}$. 
The D\&C grouping modulus is set to $N=2$, partitioning the geographic label space into $N^2=4$ disjoint groups, each associated with an independent classifier head. 
Adjacent geographic grid cells are assigned to different classifier heads, thereby mitigating perceptual aliasing between neighbouring classes within the same classifier. 
For a projected UTM coordinate $(e,n)$, the two group indices are determined by the corresponding grid-cell indices modulo $N$:
\[
\left(
\left\lfloor \frac{e}{L} \right\rfloor \bmod N,\,
\left\lfloor \frac{n}{L} \right\rfloor \bmod N
\right).
\]
Training cycles through the four predefined groups by epoch, while the classifier head of the active group is updated using the samples assigned to that group.
Each group uses an additive angular margin classifier (AAMC) with an angular margin of 0.2 and a logit scale of 100.

Before the mission sequence, the complete satellite reference dataset $\mathcal{D}_{sat}$ is used to train the initial model and to construct the static satellite exemplar memory $\mathcal{M}_{EX}$ by class-balanced herding. 
For each geographic grid cell, at most 12 representative satellite patches are retained, while classes containing fewer than 12 samples are retained in full. 
This procedure selects 91,747 unique satellite images for $\mathcal{M}_{EX}$, occupying approximately 4.93\,GB. 
The static satellite exemplar memory is separate from the bounded ER buffer and must remain accessible during continual adaptation.

The model is optimized using Adam with differential learning rates of $10^{-5}$ for the feature extractor and $10^{-3}$ for the classifier heads. 
A plateau-based scheduler decays the learning rate when the training loss plateaus. 
For methods using both memory components, each iteration uses a mixed mini-batch of 40 samples comprising
(1) 20 samples from the current-mission training set $\mathcal{D}_k^{train}$,
(2) 10 samples from the static satellite exemplar memory $\mathcal{M}_{EX}$, and
(3) 10 samples from the previous ER buffer $\mathcal{M}_{ER}^{(k-1)}$.
For the optimization objective in Sec.~\ref{sec:m_objective}, we set $\lambda_{EX}=1$ and $\lambda_{ER}=1$.
Unless otherwise stated, each classifier group has an ER-buffer capacity of $B=200$ airborne images, giving a total dynamic replay upper bound of $N^2B=800$ images. 
Neither $\mathcal{M}_{EX}$ nor $\mathcal{D}_{sat}$ is included in this 800-image replay budget. 
The current-mission training set $\mathcal{D}_k^{train}$ is also excluded from the persistent replay budget: it serves as transient adaptation input and is discarded after the buffer update except for samples selected into $\mathcal{M}_{ER}^{(k)}$.

For the DBS utility, the weighting coefficient $\lambda$ in Eq.~(\ref{eq:lambda_DBS}) is set to $1.0$. 
For Hybrid allocation, the class-wise trimming ratio is set to $\rho=0.05$, and classes containing at most two candidate samples are not trimmed. For DBS-Hybrid, this setting removes the class-wise $5\%$ tail with the largest prototype distance before coverage selection.
The number of represented classes in a group is at most 38 in the formal experiments, so the $|\mathcal{P}_{rep}^{H}|>B$ branch is not activated at $B=200$.
Unless otherwise specified, repeated experiments use three random seeds, $\{0,1,2\}$, and are reported using the mean and sample standard deviation.

For the primary C1--C3 evaluation, the distance threshold in $Acc_k(\mathcal{E};\tau)$ is set to $\tau=300\,\mathrm{m}$. 
The localization-specific analysis additionally reports Recall at $50$, $100$, $200$, and $300\,\mathrm{m}$ together with continuous position-error statistics.

\subsection{Evaluation Metrics}
\label{sec:evaluation_criteria}

To evaluate continual-learning performance across the sequence of $K=10$ discrete missions, we report standard continual-learning metrics together with C1--C3 and localization-specific measures. 
Let $R_{i,j}$ denote the test accuracy of the model on mission $j$ after completing the training of mission $i$.

\textit{Final average accuracy (FAA)} is the task-balanced mean accuracy on all task-level test splits after completing the full sequence:
\[
\text{FAA} = \frac{1}{K} \sum_{j=1}^{K} R_{K,j}.
\]

\textit{Backward transfer (BWT)} quantifies the average change in accuracy for historical tasks from their initial learning to the end of the sequence:
\[
\text{BWT} = \frac{1}{K-1} \sum_{j=1}^{K-1} (R_{K,j} - R_{j,j}).
\]
Positive BWT indicates that subsequent learning improves average performance on historical tasks, whereas negative BWT indicates average performance degradation on those tasks.

\textit{Forward transfer (FWT)} measures transfer to future tasks before they are directly learned:
\[
\text{FWT} = \frac{1}{K-1} \sum_{j=2}^{K} \left(R_{j-1,j}-R_{0,j}\right).
\]
Here, $R_{0,j}$ is the accuracy on task $j$ obtained by the shared base model before continual learning. Subtracting this reference measures the performance change attributable to prior continual learning before task $j$ is directly learned.

\textit{Average forgetting (AF)} measures the average decrease from the highest observed accuracy of each historical task to its final accuracy:
\[
\mathrm{AF}=\frac{1}{K-1}\sum_{j=1}^{K-1}\left(\max_{i\in\{j,\ldots,K\}} R_{i,j}-R_{K,j}\right).
\]
The final task is excluded because it is not followed by another mission. For each historical task $j$, the maximum includes $R_{K,j}$, so every term in the sum is non-negative. Lower AF indicates less degradation of previously attained task performance.

FAA, BWT, FWT, and AF are complemented by C1--C3 and coordinate-based geo-localization measures. We report Recall@$r$ for $r\in\{50,100,200,300\}\,\mathrm{m}$ together with median error, mean error, and root-mean-square error (RMSE). VIS and IR subsets are reported separately where their query counts permit.
For the reported C1--C3 summaries, C1 is evaluated after the final update on the held-out mission set $\mathcal{U}$, C2 averages the current-mission test accuracy immediately after each mission is learned, and C3 is evaluated after the final update on the test splits of the first $K-1$ historical missions. C3 pools the historical query sets, so missions with more test samples contribute proportionally to the score. C3 therefore excludes the current mission $s_K$, whereas the final all-mission geo-localization measures additionally include $\mathcal{D}_K^{test}$ and are not numerically equivalent to C3.

\subsection{Comparative Baselines}
\label{sec:baselines}

We compare representative continual-learning methods under a common mission-based DIL protocol for aerial VPR.
All baselines use the same feature extractor and classifier-head inference rule as our framework.
The comparison includes naive fine-tuning (FT), a DIL adaptation of Learning without Forgetting (DIL-LwF), fine-tuning with static exemplars (FT+EX), DIL adaptations of experience replay (DIL-ER), Dark Experience Replay++ (DIL-DER++), and iCaRL (DIL-iCaRL), an adapted implementation of UDIL, and uniform random replay-buffer update (Random).
The key differences are summarized in Table~\ref{tab:component_comparison}.

\begin{table*}[t]
	\centering
	\caption{\textbf{Comparison of key components across evaluated methods.} All methods use the same classifier-head inference rule. Checkmarks (\ding{51}) and crosses (\ding{55}) indicate the presence or absence of a component. ``Random'' uses a uniform random replay-buffer update over the current candidate pool. ``Memory Content'' denotes data retained between missions. KL-Div denotes Kullback--Leibler divergence, and MSE denotes mean squared error.}
	\label{tab:component_comparison}
	\resizebox{\linewidth}{!}{
		\begin{tabular}{l c c c c c c c c c}
			\toprule
			\textbf{Component} & \textbf{FT} & \textbf{DIL-LwF} & \textbf{FT+EX} & \textbf{DIL-ER} & \textbf{DIL-DER++} & \textbf{DIL-iCaRL} & \shortstack{\textbf{UDIL}\\\textbf{(adapted)}} & \textbf{Random} & \textbf{Ours} \\
			\midrule
			\textbf{Static Exemplar Memory ($\mathcal{M}_{EX}$)} & \ding{55} & \ding{55} & \ding{51} (Fixed) & \ding{55} & \ding{51} (Fixed) & \ding{51} (Fixed) & \ding{51} (Fixed) & \ding{51} (Fixed) & \ding{51} (Fixed) \\
			\textbf{ER Buffer ($\mathcal{M}_{ER}^{(k)}$)} & \ding{55} & \ding{55} & \ding{55} & Reservoir & Reservoir & Greedy Herding & Mission/Class Balanced & Uniform Random & LBS / DBS / DBS-Hybrid \\
			\textbf{Memory Content} & N/A & N/A & Images, Labels & Images, Labels & Images, Labels, Logits & Images, Labels & Images, Labels, Mission IDs, Logits, Features & Images, Labels & Images, Labels \\
			\textbf{Distillation Type} & None & KL-Div & None & None & MSE & KL-Div & KL-Div & None & None \\
			\textbf{Distillation Scope} & N/A & $\mathcal{D}_k^{train}$ & N/A & N/A & $\mathcal{M}_{ER}^{(k-1)}$ & $\mathcal{D}_k^{train} \cup \mathcal{M}_{ER}^{(k-1)} \cup \mathcal{M}_{EX}$ & $\mathcal{D}_k^{train} \cup \mathcal{M}_{ER}^{(k-1)} \cup \mathcal{M}_{EX}$ & N/A & N/A \\
			\bottomrule
		\end{tabular}
	}
\end{table*}
\subsubsection{Naive Fine-Tuning (FT)}
The model is sequentially trained on the mission sequence $\mathcal{S}$ without any memory buffer or regularization. 
This provides a fine-tuning reference for measuring forgetting across missions.

\subsubsection{DIL-LwF}
Learning without Forgetting (LwF)~\cite{lwf2016} is included as a regularization baseline adapted to the group-wise classifier architecture. At mission step $k$, each training epoch selects an active classifier group $a$, and only the current-mission training samples assigned to that group, denoted $\mathcal{D}_{k,a}^{train}\subset\mathcal{D}_k^{train}$, are used for the corresponding update. At the beginning of each active-group epoch, the current model state is copied to frozen teacher parameters $\Theta^{pre}$. For samples in $\mathcal{D}_{k,a}^{train}$, DIL-LwF combines the CE loss of the active classifier head with temperature-scaled Kullback--Leibler (KL) distillation for classifier groups $g<a$ that have already been processed in the current group cycle:
\begin{equation}
	\mathcal{L}_{LwF}^{(k,a)}=\mathbb{E}_{(x,y)\sim\mathcal{D}_{k,a}^{train}}
	\left[
	\mathcal{L}_{CE}\!\left(\mathbf{h}_{\Theta_k,a}(x),y\right)+\lambda_{LwF}
	\sum_{g<a}T^2
	D_{KL}\!\left(
	p_{\Theta^{pre},g}^{(T)}(x)
	\,\middle\|\,
	p_{\Theta_k,g}^{(T)}(x)
	\right)
	\right],
	\label{eq:dil_lwf}
\end{equation}
where $a\in\{0,1,2,3\}$ is the active classifier-group index, $\mathbf{h}_{\Theta,g}(x)$ denotes the logits produced by classifier head $g$ at model state $\Theta$, and $p_{\Theta,g}^{(T)}(x)=\operatorname{softmax}(\mathbf{h}_{\Theta,g}(x)/T)$ is its temperature-scaled class distribution. $D_{KL}$ denotes Kullback--Leibler divergence, $T$ is the distillation temperature, and $\lambda_{LwF}$ is the distillation weight. We set $T=2$ and $\lambda_{LwF}=0.5$. DIL-LwF uses neither $\mathcal{M}_{EX}$ nor an ER buffer.

\subsubsection{FT+EX (Ablation)}
FT+EX trains sequentially on $\mathcal{D}_k^{train}$ and the static satellite exemplar memory $\mathcal{M}_{EX}$ without an ER buffer.

\subsubsection{DIL-ER (Standard Experience Replay)}
To evaluate continual adaptation using dynamic replay without the static satellite exemplar memory, we implement the standard experience replay (ER) baseline~\cite{er2019Chaudhry}. Unlike our framework, DIL-ER relies on the ER buffer $\mathcal{M}_{ER}^{(k)}$ for historical replay and does not use the static satellite exemplar memory $\mathcal{M}_{EX}$. After each mission, each group buffer is updated by applying reservoir sampling sequentially to the newly observed training samples assigned to that group.

\subsubsection{DIL-DER++ (Dark Experience Replay++)}
DER++~\cite{derpp2020} is included as a replay-based continual-learning baseline adapted to the group-wise DIL architecture. The DIL-DER++ implementation uses the same static satellite exemplar memory $\mathcal{M}_{EX}$ and stores historical replay images with their ground-truth labels and classifier logits. During training, current-mission samples, static satellite exemplars, and historical replay samples form a group-wise mixed mini-batch. All labelled samples contribute to the supervised classification loss, and replay samples additionally contribute a logit-matching loss against their stored outputs.

Let $\mathcal{B}_{k,a}^{cur}$, $\mathcal{B}_{a}^{EX}$, and $\mathcal{B}_{a}^{ER}$ denote the current-mission, static satellite exemplar, and historical replay samples, respectively, for active classifier group $a$, and let $\mathcal{B}_{k,a}^{mix}=\mathcal{B}_{k,a}^{cur}\cup\mathcal{B}_{a}^{EX}\cup\mathcal{B}_{a}^{ER}$ denote the resulting mixed mini-batch. The DIL-DER++ training loss is
\begin{equation}
	\mathcal{L}_{\mathrm{DIL\text{-}DER++}}^{(k,a)}
	=
	\mathcal{L}_{CE}\!\left(\mathcal{B}_{k,a}^{mix}\right)
	+
	\beta_{\mathrm{DER}}
	\mathcal{L}_{MSE}\!\left(
	\left\{\mathbf{h}_{\Theta_k,a}(x)\right\}_{x\in\mathcal{B}_{a}^{ER}},
	\left\{\mathbf{h}^{old}(x)\right\}_{x\in\mathcal{B}_{a}^{ER}}
	\right),
	\label{eq:dil_derpp}
\end{equation}
where $\mathcal{L}_{CE}(\mathcal{B})$ denotes the mean CE loss over all labelled samples in mini-batch $\mathcal{B}$, $\mathcal{L}_{MSE}$ denotes the mean squared error (MSE) between current and stored logits over the replay mini-batch, $\mathbf{h}_{\Theta_k,a}(x)$ denotes the current logits produced by classifier head $a$, $\mathbf{h}^{old}(x)$ denotes the logits stored when replay sample $x$ entered the ER buffer, and $\beta_{\mathrm{DER}}$ controls the logit-matching loss. We set $\beta_{\mathrm{DER}}=0.5$.

\subsubsection{DIL-iCaRL (Incremental Classifier and Representation Learning)}
\label{sec:dil_icarl}
We adapt iCaRL~\cite{icarl2017}, a classical CIL method, to mission-based DIL.
The original iCaRL uses a nearest-mean-of-exemplars (NME) classifier. Here, DIL-iCaRL uses the same classifier-head inference rule as the other methods.
Regarding buffer management, DIL-iCaRL employs a \textit{greedy herding} strategy, selecting exemplars so that their mean feature representation approximates the empirical feature mean of each geographic class.
Distillation is applied to the entire group-wise mixed mini-batch, including current-mission training samples, historical replay samples, and static satellite exemplars, by matching the outputs of the frozen teacher and the current model across all spatial-group classifiers.
While the original iCaRL uses a binary cross-entropy (BCE)-based distillation objective, the DIL adaptation used here applies Kullback--Leibler (KL) divergence under the fixed geographic label space.

\subsubsection{UDIL (adapted)}
We adapt unified domain incremental learning (UDIL)~\cite{NEURIPS2023_30d046e9} to the mission-based aerial VPR setting. Each airborne mission is treated as one domain, and the model obtained after the preceding mission is frozen as the history model. Training combines supervised classification and history-model distillation on current-mission, static-exemplar, and replay samples with adversarial domain alignment and adaptive loss weights. Each of the four D\&C groups maintains an independent classifier head, adaptive coefficients, and replay memory. The group buffer is allocated across previous missions and the geographic classes observed in each mission, with deterministic backfilling when a quota cannot be filled. The supervised-contrastive term uses same-class positives from current and replay samples. Static satellite exemplars are not assigned airborne-domain labels or passed to the domain discriminator. The discriminator and adaptive weights are used only during training, while inference follows the common classifier-head rule without mission identity.

UDIL (adapted) follows the same Forward data split, initial model, mission schedule, mini-batch composition, and 800-image dynamic replay limit. Each replay entry stores the image identity, geographic label, mission ID, history logits, and 768-dimensional history features. The domain discriminator has a $768\!\to\!800\!\to\!800\!\to\!11$ architecture, of which the first $k+1$ outputs are used at mission step $k$. Its learning rate is $10^{-5}$, while the adaptive coefficients use a learning rate of $2\times10^{-3}$. The distillation temperature is 2. The encoder-alignment update uses domain-alignment and history-feature-stability coefficients of 0.1 and 10, respectively, and the adaptive-weight objective uses a generalization-bound coefficient of 5. The supervised-contrastive term has a weight of 0.1 and a temperature of 0.07 and uses negative squared Euclidean similarity. Current-mission and static-exemplar distillation is applied only when the corresponding distillation loss is below 2, while replay distillation remains weighted by the adaptive coefficients.

\subsubsection{Random (Uniform Random Replay-Buffer Update)}
\label{sec:random_baseline}

The Random baseline uses the same heterogeneous memory framework, including the static satellite exemplar memory $\mathcal{M}_{EX}$ and the bounded ER buffer $\mathcal{M}_{ER}^{(k)}$, but updates each group-wise ER buffer by uniformly sampling from the union of its previous replay samples and current-mission candidates.

\FloatBarrier

\subsection{Main Results and Analysis}
\label{sec:main_results}

\begin{table*}[t]
	\centering
	\caption{Forward three-seed continual-learning results at the primary $300\,\mathrm{m}$ distance threshold. Values are mean $\pm$ sample standard deviation. Best, second-best, and third-best mean values are shown in bold, underlined, and italic type, respectively.}
	\label{tab:table3-main-forward-3seed}
	\resizebox{\textwidth}{!}{%
		\begin{tabular}{lrrrrrrr}
			\toprule
			Method & FAA $\uparrow$ (\%) & BWT $\uparrow$ (\%) & FWT $\uparrow$ (\%) & AF $\downarrow$ (\%) & C1 $\uparrow$ (\%) & C2 $\uparrow$ (\%) & C3 $\uparrow$ (\%) \\
			\midrule
			FT & 28.97$\pm$0.58 & -1.93$\pm$0.92 & -24.06$\pm$0.12 & 11.33$\pm$0.51 & 0.80$\pm$0.13 & 30.70$\pm$0.60 & 11.95$\pm$0.80 \\
			DIL-LwF & 22.87$\pm$0.78 & -8.43$\pm$3.53 & -25.18$\pm$1.84 & 14.27$\pm$3.29 & 1.56$\pm$0.32 & 30.45$\pm$2.59 & 7.10$\pm$0.67 \\
			\addlinespace[2pt]
			FT+EX & 45.77$\pm$1.56 & -2.20$\pm$1.05 & \textbf{4.25$\pm$0.24} & 7.17$\pm$0.09 & 28.68$\pm$1.52 & \textit{47.75$\pm$1.46} & 36.95$\pm$2.07 \\
			DIL-ER & 36.69$\pm$0.77 & \textbf{2.55$\pm$1.61} & -18.62$\pm$1.07 & \textbf{4.07$\pm$1.54} & 6.17$\pm$1.16 & 34.39$\pm$1.76 & 26.77$\pm$0.70 \\
			DIL-DER++ & 46.33$\pm$1.81 & -1.27$\pm$1.57 & 2.59$\pm$1.51 & 5.69$\pm$1.51 & \textit{31.13$\pm$1.19} & 47.47$\pm$0.85 & \textit{39.26$\pm$2.06} \\
			DIL-iCaRL & 45.16$\pm$1.49 & -3.02$\pm$3.58 & 2.52$\pm$1.79 & 6.00$\pm$2.18 & 28.38$\pm$0.63 & \underline{47.88$\pm$1.82} & 37.26$\pm$1.26 \\
			UDIL (adapted) & 30.37$\pm$1.39 & -5.83$\pm$1.39 & -11.40$\pm$1.59 & 8.44$\pm$1.28 & 0.78$\pm$0.35 & 35.61$\pm$0.14 & 21.99$\pm$0.86 \\
			\addlinespace[2pt]
			Random & 42.26$\pm$0.60 & -6.41$\pm$1.04 & 2.14$\pm$1.16 & 9.29$\pm$1.57 & 27.06$\pm$1.25 & \textbf{48.03$\pm$0.84} & 33.78$\pm$0.52 \\
			LBS & \underline{46.69$\pm$2.14} & -1.01$\pm$3.72 & 2.36$\pm$1.02 & 6.17$\pm$3.78 & 27.58$\pm$1.09 & 47.61$\pm$1.63 & \underline{39.54$\pm$2.86} \\
			DBS & \textit{46.67$\pm$2.07} & \textit{-0.07$\pm$1.51} & \underline{2.61$\pm$0.75} & \textit{5.40$\pm$0.61} & \underline{32.02$\pm$1.32} & 46.74$\pm$1.82 & 39.20$\pm$1.76 \\
			DBS-Hybrid & \textbf{47.32$\pm$1.99} & \underline{1.07$\pm$1.39} & \textit{2.60$\pm$1.06} & \underline{4.37$\pm$0.23} & \textbf{32.38$\pm$0.52} & 46.35$\pm$1.64 & \textbf{40.11$\pm$1.32} \\
			\bottomrule
		\end{tabular}
	}
\end{table*}
\begin{table*}[t]
	\centering
	\caption{Forward three-seed coordinate-based geo-localization results. Values are mean $\pm$ sample standard deviation. Recall values are percentages; Recall thresholds and position errors are in metres. Best, second-best, and third-best mean values are shown in bold, underlined, and italic type, respectively.}
	\label{tab:table4-multithreshold-forward-3seed}
	\setlength{\tabcolsep}{3.2pt}
	\renewcommand{\arraystretch}{1.08}
	\resizebox{\textwidth}{!}{%
		\begin{tabular}{lrrrrrrr}
			\toprule
			Method & R@50 $\uparrow$ & R@100 $\uparrow$ & R@200 $\uparrow$ & R@300 $\uparrow$ & Median $\downarrow$ & Mean $\downarrow$ & RMSE $\downarrow$ \\
			\midrule
			\multicolumn{8}{l}{\textbf{(a) Final all-mission geo-localization}} \\
			FT & 4.09$\pm$0.13 & 14.60$\pm$0.66 & 25.08$\pm$0.12 & 27.70$\pm$0.42 & 1121.65$\pm$14.33 & 5021.56$\pm$231.16 & 9699.35$\pm$432.28 \\
			DIL-LwF & 3.88$\pm$0.12 & 14.11$\pm$0.20 & 22.38$\pm$0.19 & 24.13$\pm$0.47 & 1055.34$\pm$90.19 & 5127.02$\pm$329.03 & 9880.99$\pm$661.60 \\
			\addlinespace[2pt]
			FT+EX & 7.12$\pm$0.42 & 23.23$\pm$0.73 & 41.40$\pm$1.67 & 44.02$\pm$1.70 & 505.08$\pm$53.19 & 3844.77$\pm$185.43 & 8752.73$\pm$447.89 \\
			DIL-ER & 5.89$\pm$0.32 & 22.98$\pm$0.93 & 36.83$\pm$0.85 & 38.50$\pm$0.75 & 627.43$\pm$32.37 & \underline{2715.25$\pm$46.75} & \underline{6473.03$\pm$96.95} \\
			DIL-DER++ & 7.45$\pm$0.32 & \textit{26.55$\pm$2.35} & 43.48$\pm$1.88 & 45.85$\pm$1.87 & 405.55$\pm$57.79 & \textit{2769.10$\pm$464.41} & \textit{6942.18$\pm$942.53} \\
			DIL-iCaRL & 6.81$\pm$0.35 & 22.95$\pm$0.96 & 41.35$\pm$0.98 & 44.26$\pm$0.98 & 458.39$\pm$63.66 & 3299.14$\pm$107.62 & 8012.67$\pm$217.84 \\
			UDIL (adapted) & 4.50$\pm$0.16 & 16.50$\pm$0.34 & 31.07$\pm$0.60 & 34.18$\pm$0.87 & 798.56$\pm$18.45 & \textbf{2530.56$\pm$103.18} & \textbf{5949.47$\pm$299.63} \\
			\addlinespace[2pt]
			Random & 6.66$\pm$0.12 & 21.59$\pm$0.28 & 38.83$\pm$0.40 & 41.56$\pm$0.20 & 556.81$\pm$16.51 & 3405.20$\pm$163.70 & 8104.66$\pm$534.81 \\
			LBS & \textit{7.56$\pm$0.34} & 26.45$\pm$3.14 & \textit{43.69$\pm$2.61} & \textit{46.18$\pm$1.81} & \textit{404.24$\pm$58.00} & 2934.39$\pm$566.20 & 7301.61$\pm$1438.16 \\
			DBS & \underline{7.79$\pm$1.06} & \underline{27.40$\pm$4.68} & \underline{44.77$\pm$3.99} & \underline{47.01$\pm$2.94} & \underline{377.22$\pm$78.44} & 2833.55$\pm$285.64 & 7093.95$\pm$415.13 \\
			DBS-Hybrid & \textbf{8.07$\pm$1.05} & \textbf{28.24$\pm$4.20} & \textbf{45.64$\pm$3.56} & \textbf{47.83$\pm$2.52} & \textbf{336.12$\pm$98.09} & 2857.07$\pm$326.24 & 7275.97$\pm$585.41 \\
			\midrule
			\multicolumn{8}{l}{\textbf{(b) Generalization (C1) on held-out airborne missions}} \\
			FT & 0.01$\pm$0.02 & 0.06$\pm$0.06 & 0.62$\pm$0.20 & 0.80$\pm$0.13 & 5798.01$\pm$260.54 & 9677.03$\pm$115.49 & 14425.63$\pm$38.50 \\
			DIL-LwF & 0.23$\pm$0.19 & 0.55$\pm$0.23 & 1.38$\pm$0.22 & 1.56$\pm$0.32 & 5344.71$\pm$180.26 & 8750.47$\pm$450.78 & \underline{13209.10$\pm$579.91} \\
			\addlinespace[2pt]
			FT+EX & 5.19$\pm$0.26 & 16.48$\pm$0.79 & 27.58$\pm$1.30 & 28.68$\pm$1.52 & 4942.58$\pm$360.01 & \underline{8633.63$\pm$65.69} & \textit{13982.99$\pm$96.09} \\
			DIL-ER & 1.13$\pm$0.07 & 3.69$\pm$0.55 & 5.62$\pm$1.18 & 6.17$\pm$1.16 & \underline{4417.00$\pm$311.15} & \textbf{8245.59$\pm$601.53} & \textbf{12664.98$\pm$833.91} \\
			DIL-DER++ & \textit{5.71$\pm$0.26} & \textit{17.95$\pm$0.57} & \textit{30.22$\pm$1.07} & \textit{31.13$\pm$1.19} & \textbf{4322.60$\pm$165.65} & \textit{8725.39$\pm$498.42} & 14266.07$\pm$654.05 \\
			DIL-iCaRL & 4.92$\pm$0.31 & 15.78$\pm$0.74 & 27.54$\pm$0.56 & 28.38$\pm$0.63 & 4860.34$\pm$64.66 & 9086.35$\pm$364.98 & 14648.22$\pm$470.14 \\
			UDIL (adapted) & 0.04$\pm$0.04 & 0.10$\pm$0.08 & 0.28$\pm$0.23 & 0.78$\pm$0.35 & 4943.42$\pm$103.08 & 10253.94$\pm$667.60 & 14877.15$\pm$977.69 \\
			\addlinespace[2pt]
			Random & 4.75$\pm$0.11 & 15.07$\pm$0.87 & 26.18$\pm$1.15 & 27.06$\pm$1.25 & 4852.09$\pm$300.18 & 9007.85$\pm$166.36 & 14460.25$\pm$136.53 \\
			LBS & 4.89$\pm$0.22 & 15.68$\pm$0.37 & 26.69$\pm$1.11 & 27.58$\pm$1.09 & 4932.79$\pm$408.36 & 8977.19$\pm$791.34 & 14395.57$\pm$933.36 \\
			DBS & \underline{5.90$\pm$0.06} & \underline{18.60$\pm$1.04} & \underline{30.87$\pm$1.32} & \underline{32.02$\pm$1.32} & \textit{4443.61$\pm$282.10} & 9039.29$\pm$205.11 & 14722.06$\pm$235.22 \\
			DBS-Hybrid & \textbf{5.97$\pm$0.19} & \textbf{19.02$\pm$0.95} & \textbf{31.28$\pm$0.60} & \textbf{32.38$\pm$0.52} & 4574.81$\pm$336.03 & 9210.29$\pm$281.74 & 14957.03$\pm$299.05 \\
			\bottomrule
		\end{tabular}
	}
\end{table*}
\begin{table*}[t]
	\centering
	\caption{Forward three-seed geo-localization results for final all-mission queries and held-out Generalization (C1) queries, with each evaluation set further divided into VIS and IR subsets. Recall values are percentages and use a $300\,\mathrm{m}$ distance threshold. Best, second-best, and third-best mean values are shown in bold, underlined, and italic type, respectively.}
	\label{tab:table3-vpr-forward-3seed}
	\resizebox{\textwidth}{!}{%
		\begin{tabular}{lrrrrrr}
			\toprule
			Method & All-mission R@300 $\uparrow$ (\%) & All-mission Median $\downarrow$ (m) & All-mission VIS $\uparrow$ (\%) & All-mission IR $\uparrow$ (\%) & C1 VIS $\uparrow$ (\%) & C1 IR $\uparrow$ (\%) \\
			\midrule
			FT & 27.70$\pm$0.42 & 1121.65$\pm$14.33 & 12.79$\pm$0.27 & \textbf{65.22$\pm$0.87} & 1.03$\pm$0.17 & 0.00$\pm$0.00 \\
			DIL-LwF & 24.13$\pm$0.47 & 1055.34$\pm$90.19 & 8.05$\pm$0.76 & \underline{64.59$\pm$0.68} & 1.99$\pm$0.44 & 0.06$\pm$0.10 \\
			\addlinespace[2pt]
			FT+EX & 44.02$\pm$1.70 & 505.08$\pm$53.19 & 41.63$\pm$2.27 & 50.05$\pm$0.56 & 35.83$\pm$1.86 & 3.16$\pm$0.76 \\
			DIL-ER & 38.50$\pm$0.75 & 627.43$\pm$32.37 & 30.35$\pm$0.79 & 58.99$\pm$1.34 & 7.39$\pm$1.68 & 1.83$\pm$0.76 \\
			DIL-DER++ & 45.85$\pm$1.87 & 405.55$\pm$57.79 & \textit{44.43$\pm$2.28} & 49.41$\pm$0.95 & \textit{38.52$\pm$1.44} & \textit{4.76$\pm$0.48} \\
			DIL-iCaRL & 44.26$\pm$0.98 & 458.39$\pm$63.66 & 42.21$\pm$1.36 & 49.41$\pm$0.41 & 35.15$\pm$0.58 & 4.26$\pm$0.82 \\
			UDIL (adapted) & 34.18$\pm$0.87 & 798.56$\pm$18.45 & 22.34$\pm$0.66 & \textit{63.96$\pm$1.65} & 0.25$\pm$0.11 & 2.66$\pm$1.25 \\
			\addlinespace[2pt]
			Random & 41.56$\pm$0.20 & 556.81$\pm$16.51 & 38.29$\pm$0.59 & 49.77$\pm$1.10 & 33.57$\pm$1.61 & 3.88$\pm$0.10 \\
			LBS & \textit{46.18$\pm$1.81} & \textit{404.24$\pm$58.00} & \underline{44.79$\pm$3.30} & 49.68$\pm$2.46 & 34.17$\pm$1.64 & 4.10$\pm$0.91 \\
			DBS & \underline{47.01$\pm$2.94} & \underline{377.22$\pm$78.44} & 44.32$\pm$2.18 & 53.75$\pm$6.49 & \underline{39.61$\pm$1.63} & \underline{4.98$\pm$0.88} \\
			DBS-Hybrid & \textbf{47.83$\pm$2.52} & \textbf{336.12$\pm$98.09} & \textbf{45.40$\pm$1.62} & 53.93$\pm$6.37 & \textbf{40.00$\pm$0.68} & \textbf{5.20$\pm$0.67} \\
			\bottomrule
		\end{tabular}
	}
\end{table*}

\subsubsection{Performance under the Forward Order}


Table~\ref{tab:table3-main-forward-3seed} summarizes the three-seed continual-learning results under the Forward mission order, while Tables~\ref{tab:table4-multithreshold-forward-3seed} and~\ref{tab:table3-vpr-forward-3seed} report multi-threshold coordinate-based geo-localization performance and modality-specific results, respectively. DBS-Hybrid achieves FAA of $47.32\pm1.99\%$, BWT of $1.07\pm1.39\%$, AF of $4.37\pm0.23\%$, C1 of $32.38\pm0.52\%$, and C3 of $40.11\pm1.32\%$. Relative to Random, these results improve FAA by $5.06$ percentage points, BWT by $7.48$ points, C1 by $5.32$ points, and C3 by $6.33$ points, while decreasing AF by $4.92$ points. Random retains a higher mean C2 ($48.03\%$ versus $46.35\%$), while FT+EX obtains the highest FWT ($4.25\pm0.24\%$). DIL-ER achieves the highest BWT ($2.55\pm1.61\%$) and the lowest AF ($4.07\pm1.54\%$), with FAA of $36.69\pm0.77\%$, C1 of $6.17\pm1.16\%$, and C3 of $26.77\pm0.70\%$. DIL-LwF obtains FAA of $22.87\pm0.78\%$, C1 of $1.56\pm0.32\%$, and C3 of $7.10\pm0.67\%$. Adapted UDIL obtains FAA of $30.37\pm1.39\%$, C1 of $0.78\pm0.35\%$, and C3 of $21.99\pm0.86\%$. These results characterize immediate adaptation (C2), knowledge retention (C3), and generalization (C1), and show that DBS-Hybrid achieves the highest FAA, C1, and C3 under the Forward order.

\begin{figure}[!t]
	\centering
	\includegraphics[width=0.6\linewidth]{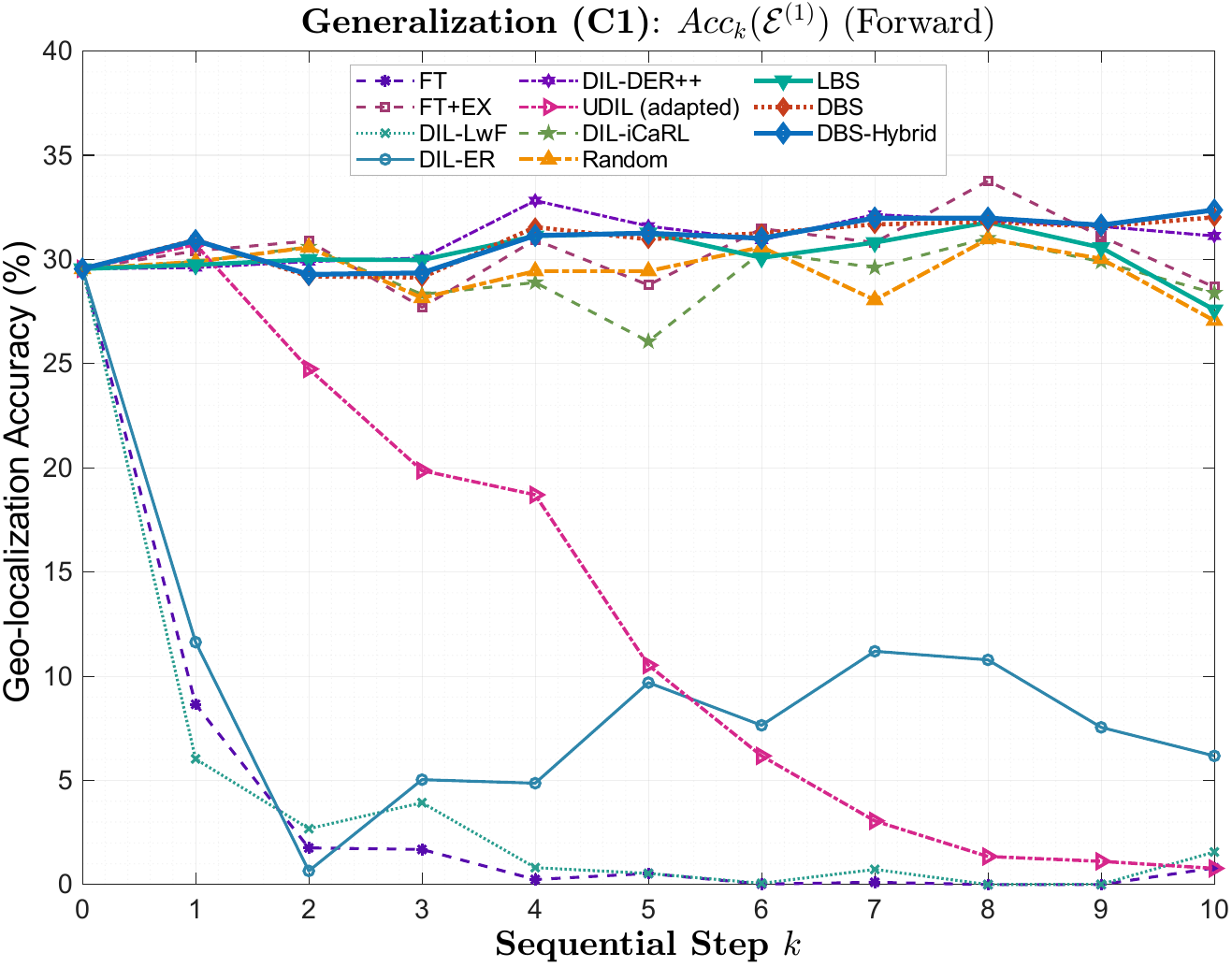}\\
	\vspace{6pt}
	\includegraphics[width=0.6\linewidth]{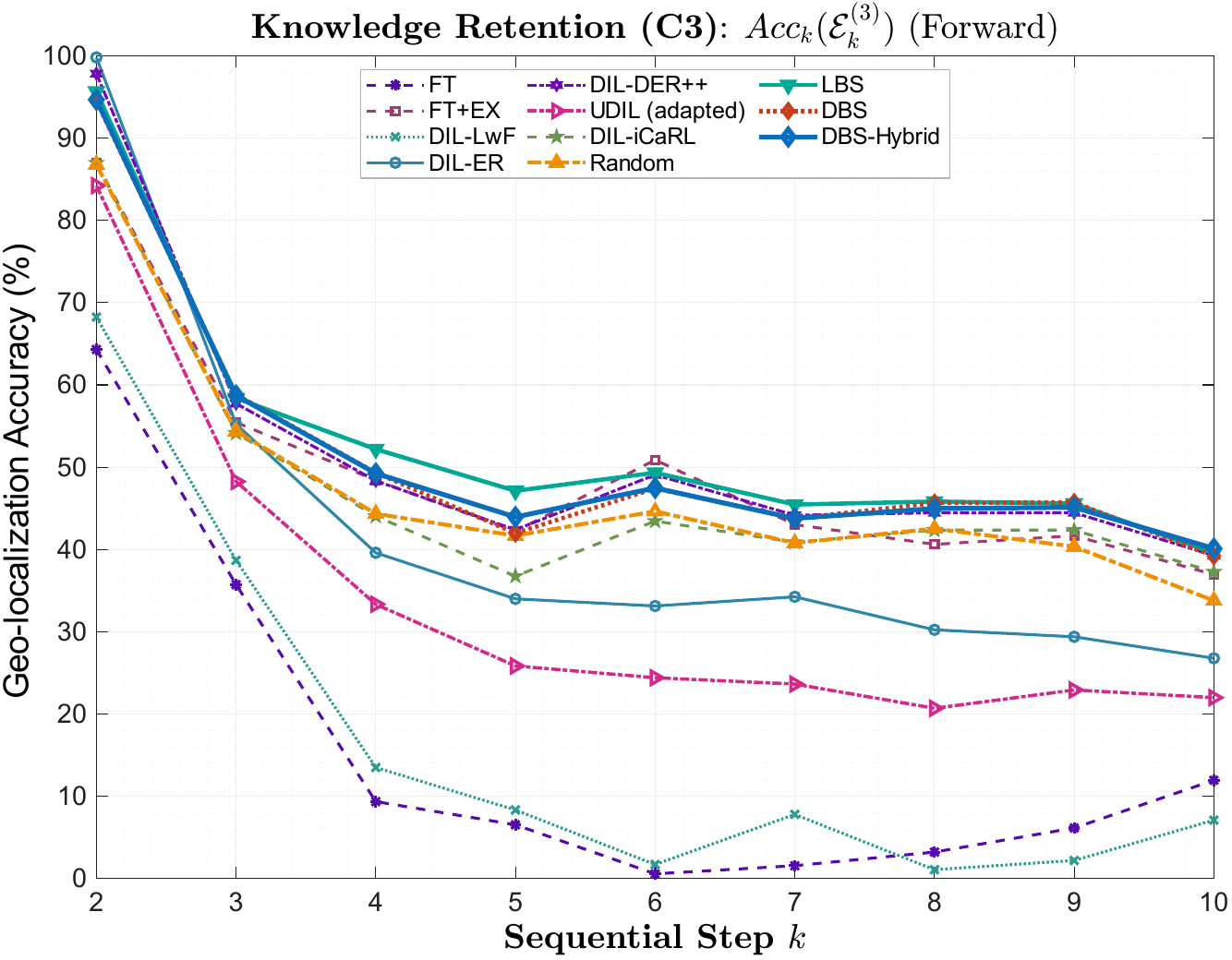}
	\caption{\textbf{Metric trajectories under the Forward mission order.} Curves show the three-seed mean at each step and follow the method order used in Table~\ref{tab:table3-main-forward-3seed}: FT, DIL-LwF, FT+EX, DIL-ER, DIL-DER++, DIL-iCaRL, UDIL (adapted), Random, LBS, DBS, and DBS-Hybrid. \textbf{(Top)} Generalization (C1) on held-out airborne missions, $Acc_k(\mathcal{E}^{(1)})$; $k=0$ denotes the initial model before continual adaptation. At $k=10$, DBS-Hybrid records the highest mean C1. \textbf{(Bottom)} Knowledge retention (C3), $Acc_k(\mathcal{E}_k^{(3)})$. At $k=10$, DBS-Hybrid records the highest mean C3.}
	\label{fig:forward_metrics}
\end{figure}

Across all four distance thresholds in Table~\ref{tab:table4-multithreshold-forward-3seed}, DBS-Hybrid achieves the highest mean recall for both final all-mission geo-localization and generalization (C1). Its final all-mission recalls at 50, 100, 200, and $300\,\mathrm{m}$ are $8.07\%$, $28.24\%$, $45.64\%$, and $47.83\%$, respectively, compared with $6.66\%$, $21.59\%$, $38.83\%$, and $41.56\%$ for Random. On the held-out airborne missions used for C1, the corresponding recalls are $5.97\%$, $19.02\%$, $31.28\%$, and $32.38\%$ for DBS-Hybrid, compared with $4.75\%$, $15.07\%$, $26.18\%$, and $27.06\%$ for Random. Because each predicted class is converted to the centre of a $200\,\mathrm{m}$ grid cell, Recall@$50\,\mathrm{m}$ measures localization within a radius smaller than the cell side length. DBS-Hybrid remains ahead of Random at every reported threshold.

The continuous position-error statistics provide a complementary view. DBS-Hybrid yields the lowest final all-mission median error ($336.12\pm98.09\,\mathrm{m}$), whereas adapted UDIL achieves the lowest mean error ($2530.56\pm103.18\,\mathrm{m}$) and RMSE ($5949.47\pm299.63\,\mathrm{m}$). Adapted UDIL obtains final all-mission Recall@$300\,\mathrm{m}$ of $34.18\pm0.87\%$ and a median error of $798.56\pm18.45\,\mathrm{m}$. For C1, DIL-DER++ gives the lowest median error ($4322.60\pm165.65\,\mathrm{m}$), while DIL-ER gives the lowest mean error and RMSE. These statistics characterize the central and long-range parts of the localization-error distribution. The modality-specific results in Table~\ref{tab:table3-vpr-forward-3seed} show different results for VIS and IR queries. DBS-Hybrid achieves the highest final all-mission VIS recall ($45.40\pm1.62\%$) and the highest C1 recall for both VIS ($40.00\pm0.68\%$) and IR ($5.20\pm0.67\%$). Its final all-mission IR recall is $53.93\pm6.37\%$; FT, DIL-LwF, adapted UDIL ($63.96\pm1.65\%$), and DIL-ER obtain higher values, while Random, LBS, and DBS obtain lower values. The Forward order ends with the 500-image IR mission \texttt{TD-13-seq}, so the final IR result reflects both sensor modality and the effect of the last training mission.


\subsubsection{Sensitivity to Mission Order}
\label{sec:robustness}

The primary Forward order provides the complete baseline comparison. To assess sensitivity to mission order, we evaluate DIL-DER++, DIL-iCaRL, Random, LBS, DBS, and DBS-Hybrid under the four named mission orders defined in Sec.~\ref{sec:curriculum} and five additional fixed random mission orders. The remaining baselines are reported under the Forward order.

\begin{table*}[t]
	\centering
	\caption{FAA / BWT (\%) for the six methods included in the mission-order analysis across four named mission orders. Values are mean $\pm$ sample standard deviation over three seeds. Within each metric and order, best, second-best, and third-best mean values are shown in bold, underlined, and italic type, respectively.}
	\label{tab:table4-named-orders-final-bwt}
	\resizebox{\textwidth}{!}{%
		\begin{tabular}{lrrrr}
			\toprule
			Method & Forward & Backward & Pressure & Robust \\
			\midrule
			DIL-DER++ & 46.33$\pm$1.81 / -1.27$\pm$1.57 & \textbf{49.58$\pm$1.09} / \textbf{2.91$\pm$0.88} & \underline{50.99$\pm$0.99} / 1.46$\pm$1.72 & 46.34$\pm$1.59 / \textit{0.25$\pm$0.71} \\
			DIL-iCaRL & 45.16$\pm$1.49 / -3.02$\pm$3.58 & \underline{48.84$\pm$1.87} / -2.38$\pm$0.74 & 50.75$\pm$2.50 / 1.67$\pm$3.10 & 46.13$\pm$2.25 / -1.53$\pm$2.49 \\
			Random & 42.26$\pm$0.60 / -6.41$\pm$1.04 & 47.95$\pm$2.79 / -0.81$\pm$0.37 & \textbf{51.33$\pm$0.79} / \textbf{4.05$\pm$2.32} & 47.06$\pm$3.42 / -0.16$\pm$2.87 \\
			LBS & \underline{46.69$\pm$2.14} / \textit{-1.01$\pm$3.72} & \textit{48.68$\pm$2.37} / \underline{1.83$\pm$2.10} & 49.19$\pm$2.61 / 0.20$\pm$0.17 & \textbf{49.16$\pm$3.92} / -0.11$\pm$3.45 \\
			DBS & \textit{46.67$\pm$2.07} / \underline{-0.07$\pm$1.51} & 48.34$\pm$0.61 / \textit{-0.46$\pm$1.04} & 49.96$\pm$0.16 / \textit{2.05$\pm$0.17} & \textit{48.37$\pm$1.78} / \underline{0.69$\pm$1.15} \\
			DBS-Hybrid & \textbf{47.32$\pm$1.99} / \textbf{1.07$\pm$1.39} & 48.18$\pm$0.59 / -0.53$\pm$0.77 & \textit{50.81$\pm$0.77} / \underline{2.91$\pm$2.02} & \underline{48.61$\pm$1.81} / \textbf{1.16$\pm$1.79} \\
			\bottomrule
		\end{tabular}
	}
\end{table*}

\begin{table*}[t]
	\centering
	\caption{C1 / C2 / C3 (\%) for the six methods included in the mission-order analysis across four named mission orders. Values are mean $\pm$ sample standard deviation over three seeds. Within each criterion and order, best, second-best, and third-best mean values are shown in bold, underlined, and italic type, respectively.}
	\label{tab:table5-named-orders-c1-c2-c3}
	\resizebox{\textwidth}{!}{%
		\begin{tabular}{lrrrr}
			\toprule
			Method & Forward & Backward & Pressure & Robust \\
			\midrule
			DIL-DER++ & \textit{31.13$\pm$1.19} / 47.47$\pm$0.85 / \textit{39.26$\pm$2.06} & \textbf{32.02$\pm$0.49} / 46.95$\pm$0.41 / \underline{44.33$\pm$2.64} & 32.51$\pm$1.40 / \textbf{49.67$\pm$1.46} / \textit{51.25$\pm$1.25} & \textit{31.76$\pm$0.40} / 46.12$\pm$1.79 / 44.43$\pm$1.79 \\
			DIL-iCaRL & 28.38$\pm$0.63 / \underline{47.88$\pm$1.82} / 37.26$\pm$1.26 & 30.87$\pm$1.24 / \textbf{50.99$\pm$2.21} / \textbf{44.77$\pm$1.81} & \underline{32.79$\pm$1.99} / \underline{49.24$\pm$0.92} / 51.19$\pm$2.40 & 31.37$\pm$0.86 / 47.50$\pm$0.64 / 44.11$\pm$2.25 \\
			Random & 27.06$\pm$1.25 / \textbf{48.03$\pm$0.84} / 33.78$\pm$0.52 & 28.29$\pm$1.81 / \textit{48.68$\pm$2.66} / 44.09$\pm$2.38 & \textit{32.69$\pm$1.97} / 47.69$\pm$2.43 / \textbf{51.90$\pm$0.69} & 31.13$\pm$1.47 / 47.21$\pm$1.40 / 45.97$\pm$4.72 \\
			LBS & 27.58$\pm$1.09 / \textit{47.61$\pm$1.63} / \underline{39.54$\pm$2.86} & 29.94$\pm$0.78 / 47.03$\pm$2.25 / \textit{44.21$\pm$2.28} & 31.07$\pm$0.34 / \textit{49.01$\pm$2.64} / 48.94$\pm$3.99 & 31.71$\pm$0.23 / \textbf{49.26$\pm$0.87} / \textbf{48.81$\pm$5.10} \\
			DBS & \underline{32.02$\pm$1.32} / 46.74$\pm$1.82 / 39.20$\pm$1.76 & \textit{31.20$\pm$0.34} / \underline{48.76$\pm$0.34} / 43.88$\pm$0.70 & 32.67$\pm$1.00 / 48.11$\pm$0.26 / 50.90$\pm$1.19 & \underline{32.58$\pm$0.28} / \underline{47.75$\pm$2.43} / \textit{47.18$\pm$3.20} \\
			DBS-Hybrid & \textbf{32.38$\pm$0.52} / 46.35$\pm$1.64 / \textbf{40.11$\pm$1.32} & \underline{31.84$\pm$0.80} / 48.66$\pm$0.54 / 43.79$\pm$0.55 & \textbf{32.85$\pm$1.21} / 48.19$\pm$1.78 / \underline{51.56$\pm$0.50} & \textbf{33.01$\pm$1.05} / \textit{47.57$\pm$3.10} / \underline{47.29$\pm$3.35} \\
			\bottomrule
		\end{tabular}
	}
\end{table*}

The named-order results in Tables~\ref{tab:table4-named-orders-final-bwt} and~\ref{tab:table5-named-orders-c1-c2-c3} show that mission order affects the balance between adaptation and retention. DBS-Hybrid achieves the highest FAA, BWT, C1, and C3 under Forward; its C1 ranks second under Backward and first under Pressure and Robust, and its BWT is also highest under Robust. These results show strong joint retention and generalization under Forward and consistently high held-out generalization across the other named orders.

\begin{table*}[t]
	\centering
	\caption{Across-order summary for the six methods included in the mission-order analysis over five additional fixed random mission orders. Values are mean $\pm$ sample standard deviation across the five orders; the variability reflects mission-order variation rather than repeated-seed variation. Best, second-best, and third-best mean values are shown in bold, underlined, and italic type, respectively.}
	\label{tab:random-order-summary}
	\begin{tabular}{lrrrrr}
		\toprule
		Method & FAA $\uparrow$ (\%) & BWT $\uparrow$ (\%) & C1 $\uparrow$ (\%) & C2 $\uparrow$ (\%) & C3 $\uparrow$ (\%) \\
		\midrule
		DIL-DER++ & \textbf{50.39$\pm$2.03} & \textbf{2.76$\pm$2.70} & \textbf{32.22$\pm$1.80} & 47.91$\pm$0.95 & \textbf{49.39$\pm$5.39} \\
		DIL-iCaRL & 47.21$\pm$3.49 & -2.95$\pm$3.57 & \textit{31.39$\pm$0.76} & \textbf{49.87$\pm$1.28} & 46.17$\pm$4.09 \\
		Random & 48.22$\pm$2.40 & -0.69$\pm$3.29 & 29.41$\pm$2.83 & \textit{48.84$\pm$0.91} & 47.64$\pm$7.09 \\
		LBS & 47.91$\pm$3.24 & -1.85$\pm$1.48 & 29.14$\pm$1.41 & \underline{49.58$\pm$2.04} & 46.93$\pm$7.48 \\
		DBS & \textit{48.38$\pm$3.19} & \textit{-0.47$\pm$4.04} & 31.23$\pm$2.04 & 48.80$\pm$4.01 & \textit{48.32$\pm$8.01} \\
		DBS-Hybrid & \underline{48.66$\pm$2.16} & \underline{0.61$\pm$2.92} & \underline{32.07$\pm$2.20} & 48.11$\pm$3.04 & \underline{48.66$\pm$6.96} \\
		\bottomrule
	\end{tabular}
\end{table*}

The five additional fixed random mission orders provide a complementary assessment of order sensitivity.
As summarized in Table~\ref{tab:random-order-summary}, DIL-DER++ obtains the highest across-order mean FAA ($50.39\pm2.03\%$), BWT ($2.76\pm2.70\%$), C1 ($32.22\pm1.80\%$), and C3 ($49.39\pm5.39\%$), whereas DBS-Hybrid ranks second on these four measures with $48.66\pm2.16\%$, $0.61\pm2.92\%$, $32.07\pm2.20\%$, and $48.66\pm6.96\%$, respectively.
The differences between DIL-DER++ and DBS-Hybrid are comparatively small for C1 ($0.15$ percentage points) and C3 ($0.73$ percentage points), while their differences are larger for FAA and BWT.
Each standard deviation in Table~\ref{tab:random-order-summary} describes variability across the five fixed mission orders rather than variability across repeated random seeds.

The detailed per-order results in \ref{sec:appendix_random_orders} provide a complementary view of this comparison.
Across the five additional random mission orders, DBS-Hybrid achieves the highest mean FWT ($1.90\%$) and the second-lowest AF ($3.76\%$), whereas DIL-DER++ achieves the lowest mean AF ($3.25\%$).
Together, the primary and additional-order results show that DBS-Hybrid achieves the highest FAA, C1, and C3 with positive BWT under the Forward order and ranks second on the four corresponding means across the five additional random orders.
The selection-criterion and allocation experiments in Sec.~\ref{sec:ablation} examine the DBS-Hybrid design under the primary Forward order.

\FloatBarrier

\subsection{Additional Experiments}

The primary experiments use DINOv2+GeM on the mission-based aerial benchmark. We conduct two additional experiments: DINOv2+SALAD under the Forward mission order and an altitude-based sequential evaluation on the public SUES-200 dataset.

\subsubsection{DINOv2+SALAD}

We evaluate Random and DBS-Hybrid using DINOv2+SALAD under the Forward mission order with the same three random seeds. This comparison changes the aggregation method from GeM to SALAD while retaining the DINOv2 backbone, dual-memory setting, mission order, and replay budget.

\begin{table*}[t]
	\centering
	\caption{DINOv2+SALAD results under the Forward mission order. Values are mean $\pm$ sample standard deviation over three seeds. Continual-learning and recall metrics are percentages, and median position error is reported in metres. Best and second-best mean values are shown in bold and underlined type, respectively.}
	\label{tab:supp-salad-interaction}
	\setlength{\tabcolsep}{3.2pt}
	\resizebox{\textwidth}{!}{%
		\begin{tabular}{lrrrrrrrrr}
			\toprule
			Method & FAA $\uparrow$ (\%) & BWT $\uparrow$ (\%) & FWT $\uparrow$ (\%) & AF $\downarrow$ (\%) & C1 $\uparrow$ (\%) & C2 $\uparrow$ (\%) & C3 $\uparrow$ (\%) & All-mission R@300 $\uparrow$ (\%) & Median $\downarrow$ (m) \\
			\midrule
			Random & \underline{42.025$\pm$3.696} & \underline{-11.233$\pm$4.128} & \textbf{7.179$\pm$0.788} & \underline{13.119$\pm$4.400} & \underline{27.305$\pm$0.729} & \textbf{52.135$\pm$1.393} & \underline{35.425$\pm$3.998} & \underline{43.151$\pm$3.433} & \underline{491.931$\pm$102.459} \\
			DBS-Hybrid & \textbf{46.222$\pm$2.127} & \textbf{-5.433$\pm$1.317} & \underline{6.730$\pm$1.244} & \textbf{9.292$\pm$1.428} & \textbf{33.333$\pm$1.370} & \underline{51.112$\pm$1.014} & \textbf{39.132$\pm$2.407} & \textbf{45.695$\pm$1.987} & \textbf{425.587$\pm$54.156} \\
			\bottomrule
		\end{tabular}%
	}
\end{table*}

With DINOv2+SALAD, DBS-Hybrid exceeds Random by $4.197$ percentage points in FAA, $5.800$ points in BWT, $6.028$ points in C1, and $3.707$ points in C3, while reducing AF by $3.827$ points. Final all-mission Recall@$300\,\mathrm{m}$ increases from $43.151\%$ to $45.695\%$, and the median position error decreases from $491.931$ to $425.587\,\mathrm{m}$. Random retains slightly higher FWT and C2. DBS-Hybrid continues to achieve higher FAA, BWT, C1, C3, and final all-mission Recall@$300\,\mathrm{m}$, together with lower AF and median position error, after GeM is replaced with SALAD.

\subsubsection{SUES-200}

SUES-200 provides drone observations at four flight heights. We organize the observations acquired at 150, 200, 250, and $300\,\mathrm{m}$ as four sequential pseudo-missions. For each training location and altitude, the first 40 drone images are used for learning and the remaining 10 images form the corresponding task-level test split. After the four sequential updates, the final model is evaluated separately on the 80 held-out locations in the standard SUES-200 drone-to-satellite retrieval protocol.

Let $\widetilde{R}_{i,j}$ denote the test accuracy on pseudo-mission $j$ after learning pseudo-mission $i$. In addition to FAA, BWT, and AF, we report the final performance retained on the first three pseudo-missions as
\begin{equation}
	\mathrm{Old\text{-}3\ Final}
	=
	\frac{1}{3}
	\sum_{j=1}^{3}
	\widetilde{R}_{4,j}.
	\label{eq:sues_old3}
\end{equation}
AF follows the same definition as in the primary continual-learning evaluation with $K=4$. The pseudo-mission evaluation compares Random, LBS, and DBS-Hybrid, while the controlled decomposition of utility and allocation is examined separately in Sec.~\ref{sec:ablation}.

\begin{table*}[t]
	\centering
	\caption{Results on SUES-200. Panel (a) reports continual-learning performance on the altitude-based pseudo-mission sequence, and Panel (b) reports drone-to-satellite retrieval on the 80 held-out locations in the standard SUES-200 protocol after the four pseudo-missions. Values are mean $\pm$ sample standard deviation over three seeds. Best and second-best mean values are shown in bold and underlined type, respectively.}
	\label{tab:sues200-validation}
	\setlength{\tabcolsep}{4pt}
	\renewcommand{\arraystretch}{1.08}
	\resizebox{\textwidth}{!}{%
		\begin{tabular}{lrrrrr}
			\toprule
			\multicolumn{6}{l}{\textbf{(a) Altitude-based pseudo-mission continual learning}} \\
			\midrule
			Method &
			FAA $\uparrow$ (\%) &
			BWT $\uparrow$ (\%) &
			AF $\downarrow$ (\%) &
			\multicolumn{2}{c}{Old-3 Final $\uparrow$ (\%)} \\
			\midrule
			Random &
			\underline{98.979$\pm$0.116} &
			\underline{0.398$\pm$0.363} &
			\underline{0.124$\pm$0.056} &
			\multicolumn{2}{c}{\underline{98.972$\pm$0.100}} \\
			LBS &
			98.972$\pm$0.167 &
			0.306$\pm$0.558 &
			0.209$\pm$0.191 &
			\multicolumn{2}{c}{98.954$\pm$0.170} \\
			DBS-Hybrid &
			\textbf{99.104$\pm$0.075} &
			\textbf{0.546$\pm$0.485} &
			\textbf{0.012$\pm$0.021} &
			\multicolumn{2}{c}{\textbf{99.093$\pm$0.085}} \\
			\midrule
			\multicolumn{6}{l}{\textbf{(b) Standard SUES-200 unseen-location retrieval}} \\
			\midrule
			Method &
			R@1 $\uparrow$ (\%) &
			R@5 $\uparrow$ (\%) &
			R@10 $\uparrow$ (\%) &
			mAP $\uparrow$ (\%) &
			Mean rank $\downarrow$ \\
			\midrule
			Random &
			\textbf{91.388$\pm$1.498} &
			\underline{98.465$\pm$0.672} &
			\textbf{99.363$\pm$0.485} &
			\textbf{94.632$\pm$0.640} &
			\textbf{1.270$\pm$0.085} \\
			LBS &
			89.817$\pm$2.068 &
			\textbf{98.494$\pm$0.498} &
			\underline{99.323$\pm$0.367} &
			93.794$\pm$1.383 &
			\underline{1.293$\pm$0.106} \\
			DBS-Hybrid &
			\underline{91.023$\pm$1.464} &
			97.975$\pm$0.629 &
			98.804$\pm$0.996 &
			\underline{94.172$\pm$1.066} &
			1.429$\pm$0.273 \\
			\bottomrule
		\end{tabular}%
	}
\end{table*}

Under the altitude-based pseudo-mission protocol, DBS-Hybrid achieves the highest FAA ($99.104\pm0.075\%$), BWT ($0.546\pm0.485\%$), and Old-3 Final ($99.093\pm0.085\%$), together with the lowest AF ($0.012\pm0.021\%$), among the three evaluated methods (Table~\ref{tab:sues200-validation}(a)). DBS-Hybrid therefore gives the strongest continual-retention results on this four-step altitude sequence.

On the 80 held-out locations, Random achieves the highest mean R@1 ($91.388\pm1.498\%$), mAP ($94.632\pm0.640\%$), and mean rank ($1.270\pm0.085$), while DBS-Hybrid ranks second in R@1 and mAP (Table~\ref{tab:sues200-validation}(b)). The two panels distinguish continual retention over the altitude sequence from final static retrieval on held-out locations.

\FloatBarrier

\subsection{Ablation Study}
\label{sec:ablation}

\subsubsection{Utility and Allocation Analysis}

We first examine the allocation of a fixed scalar utility by comparing three operators: (i) \textit{Global-within-group}, which retains the highest-utility samples in each classifier group without an explicit class constraint; (ii) \textit{Round-Robin}, which distributes replay capacity approximately evenly across represented classes; and (iii) \textit{Min-Guar}, which prioritizes one high-utility representative from each represented class whenever the budget permits and allocates the remaining capacity according to utility. Table~\ref{tab:table6-allocation} applies these three operators separately to the LBS and DBS utilities under the Forward mission order.

\begin{table*}[t]
	\centering
	\caption{Three-seed scalar-score allocation analysis under the Forward mission order with $B=200$ per classifier group. Global-within-group, Round-Robin, and Min-Guar operate on the same scalar utility within each LBS or DBS block. Best, second-best, and third-best mean values are shown in bold, underlined, and italic type, respectively.}
	\label{tab:table6-allocation}
	\resizebox{\textwidth}{!}{%
		\begin{tabular}{llrrrrrrr}
			\toprule
			Utility & Allocation & FAA $\uparrow$ (\%) & BWT $\uparrow$ (\%) & FWT $\uparrow$ (\%) & AF $\downarrow$ (\%) & C1 $\uparrow$ (\%) & C2 $\uparrow$ (\%) & C3 $\uparrow$ (\%) \\
			\midrule
			LBS & Global-within-group & \textit{46.23$\pm$4.09} & -2.01$\pm$4.72 & 2.27$\pm$2.22 & 6.72$\pm$4.26 & \textit{31.32$\pm$1.35} & \underline{48.04$\pm$0.47} & \underline{39.35$\pm$2.55} \\
			LBS & Round-Robin & 45.95$\pm$2.97 & -3.54$\pm$2.30 & \textit{2.40$\pm$1.16} & 7.38$\pm$1.34 & \underline{31.53$\pm$1.84} & \textbf{49.14$\pm$0.95} & 38.24$\pm$2.59 \\
			LBS & Min-Guar & \textbf{46.69$\pm$2.14} & \underline{-1.01$\pm$3.72} & 2.36$\pm$1.02 & \textit{6.17$\pm$3.78} & 27.58$\pm$1.09 & \textit{47.61$\pm$1.63} & \textbf{39.54$\pm$2.86} \\
			DBS & Global-within-group & 44.65$\pm$0.54 & -2.52$\pm$3.75 & 1.52$\pm$0.98 & \underline{5.74$\pm$1.63} & 30.74$\pm$1.56 & 46.92$\pm$3.72 & 37.45$\pm$1.60 \\
			DBS & Round-Robin & 45.12$\pm$0.96 & \textit{-1.72$\pm$1.71} & \textbf{3.46$\pm$1.86} & 6.61$\pm$1.51 & 30.85$\pm$0.49 & 46.66$\pm$0.89 & 38.24$\pm$1.46 \\
			DBS & Min-Guar & \underline{46.67$\pm$2.07} & \textbf{-0.07$\pm$1.51} & \underline{2.61$\pm$0.75} & \textbf{5.40$\pm$0.61} & \textbf{32.02$\pm$1.32} & 46.74$\pm$1.82 & \textit{39.20$\pm$1.76} \\
			\bottomrule
		\end{tabular}
	}
\end{table*}

Within the DBS block, Min-Guar gives the highest FAA, BWT, C1, and C3 and the lowest AF among the three scalar-score allocation operators, whereas Round-Robin gives the highest FWT. Within the LBS block, Min-Guar gives the highest FAA, BWT, and C3 and the lowest AF, while Round-Robin gives the highest FWT, C1, and C2. Min-Guar therefore provides the strongest retention-oriented allocation for both scalar utilities and serves as the representative-first baseline for Hybrid.

\begin{table*}[t]
	\centering
	\caption{Three-seed component analysis of DBS-Hybrid under the Forward mission order. Representative-only and $k$-center-only are structurally reduced selectors, whereas DBS-Hybrid uses the complete representative, trimming, coverage, and backfill sequence. All metrics are percentages. Best and second-best mean values are shown in bold and underlined type, respectively.}
	\label{tab:table6-component-ablation}
	\resizebox{\textwidth}{!}{%
		\begin{tabular}{lrrrrrrrr}
			\toprule
			Selection & FAA $\uparrow$ (\%) & BWT $\uparrow$ (\%) & FWT $\uparrow$ (\%) & AF $\downarrow$ (\%) & C1 $\uparrow$ (\%) & C2 $\uparrow$ (\%) & C3 $\uparrow$ (\%) & All-mission R@300 $\uparrow$ (\%) \\
			\midrule
			Representative-only & 46.35$\pm$1.24 & \underline{0.44$\pm$2.18} & \textbf{3.76$\pm$1.18} & 5.19$\pm$1.88 & \underline{31.60$\pm$1.46} & 45.95$\pm$0.91 & \underline{39.48$\pm$1.62} & \underline{46.29$\pm$1.39} \\
			$k$-center-only & \underline{46.42$\pm$0.67} & 0.01$\pm$1.75 & \underline{2.77$\pm$0.84} & \underline{5.09$\pm$1.30} & 31.03$\pm$0.08 & \textbf{46.41$\pm$0.90} & 39.23$\pm$1.39 & 46.05$\pm$1.19 \\
			DBS-Hybrid & \textbf{47.32$\pm$1.99} & \textbf{1.07$\pm$1.39} & 2.60$\pm$1.06 & \textbf{4.37$\pm$0.23} & \textbf{32.38$\pm$0.52} & \underline{46.35$\pm$1.64} & \textbf{40.11$\pm$1.32} & \textbf{47.83$\pm$2.52} \\
			\bottomrule
		\end{tabular}
	}
\end{table*}

Table~\ref{tab:table6-component-ablation} compares DBS-Hybrid with representative-only and $k$-center-only selectors. DBS-Hybrid gives the highest mean FAA, BWT, C1, C3, and final all-mission Recall@$300\,\mathrm{m}$ and the lowest AF, whereas representative-only gives the highest FWT and $k$-center-only gives the highest C2. Thus, the complete representative--trimming--coverage--backfill sequence improves mean FAA, BWT, C1, C3, and final all-mission Recall@$300\,\mathrm{m}$ and reduces AF relative to either reduced selector.

To compare the trimming criteria under the same Hybrid allocation, we construct LBS-Hybrid as a diagnostic variant of DBS-Hybrid. LBS-Hybrid replaces the prototype-similarity trimming score $q_H$ in Eq.~(\ref{eq:hybrid_score}) with the loss-based score $\sigma_{LBS}$, while retaining the same prototype-nearest representatives, cosine $k$-center coverage, and deterministic backfilling.

\begin{table*}[t]
	\centering
	\caption{Three-seed analysis of selection criterion and allocation under the Forward DINOv2+GeM setting with $B=200$ per classifier group. LBS and DBS use scalar utility with Min-Guar. LBS-Hybrid is a diagnostic variant that replaces the prototype-similarity trimming score of DBS-Hybrid with $\sigma_{LBS}$ while retaining the same Hybrid allocation. Best, second-best, and third-best mean values are shown in bold, underlined, and italic type, respectively.}
	\label{tab:utility-operator-interaction}
	\resizebox{\textwidth}{!}{%
		\begin{tabular}{llrrrrrrrr}
			\toprule
			Selection criterion & Allocation & FAA $\uparrow$ (\%) & BWT $\uparrow$ (\%) & FWT $\uparrow$ (\%) & AF $\downarrow$ (\%) & C1 $\uparrow$ (\%) & C2 $\uparrow$ (\%) & C3 $\uparrow$ (\%) & All-mission R@300 $\uparrow$ (\%) \\
			\midrule
			Difficulty & Min-Guar (LBS) & \underline{46.69$\pm$2.14} & \textit{-1.01$\pm$3.72} & \textit{2.36$\pm$1.02} & \textit{6.17$\pm$3.78} & 27.58$\pm$1.09 & \textbf{47.61$\pm$1.63} & \underline{39.54$\pm$2.86} & \textit{46.18$\pm$1.81} \\
			Difficulty & Hybrid (LBS-Hybrid) & 44.42$\pm$0.95 & -1.76$\pm$1.77 & 0.41$\pm$1.81 & 6.39$\pm$1.10 & \textit{31.22$\pm$1.40} & 46.01$\pm$0.66 & 37.10$\pm$0.55 & 44.20$\pm$0.35 \\
			Diversity & Min-Guar (DBS) & \textit{46.67$\pm$2.07} & \underline{-0.07$\pm$1.51} & \textbf{2.61$\pm$0.75} & \underline{5.40$\pm$0.61} & \underline{32.02$\pm$1.32} & \underline{46.74$\pm$1.82} & \textit{39.20$\pm$1.76} & \underline{47.01$\pm$2.94} \\
			Diversity & Hybrid (DBS-Hybrid) & \textbf{47.32$\pm$1.99} & \textbf{1.07$\pm$1.39} & \underline{2.60$\pm$1.06} & \textbf{4.37$\pm$0.23} & \textbf{32.38$\pm$0.52} & \textit{46.35$\pm$1.64} & \textbf{40.11$\pm$1.32} & \textbf{47.83$\pm$2.52} \\
			\bottomrule
		\end{tabular}%
	}
\end{table*}

Table~\ref{tab:utility-operator-interaction} provides a diagnostic comparison of difficulty- and diversity-oriented replay selection under Min-Guar and the corresponding Hybrid configurations. Under Min-Guar, DBS exceeds LBS by $0.94$ percentage points in BWT, $0.25$ points in FWT, $4.44$ points in C1, and $0.83$ points in final all-mission Recall@$300\,\mathrm{m}$, while decreasing AF by $0.77$ points. LBS is higher by $0.02$ points in FAA, $0.87$ points in C2, and $0.34$ points in C3.

Under Hybrid allocation, DBS-Hybrid exceeds LBS-Hybrid by approximately $2.90$ percentage points in FAA, $2.83$ points in BWT, $2.19$ points in FWT, $1.16$ points in C1, $0.34$ points in C2, $3.01$ points in C3, and $3.63$ points in final all-mission Recall@$300\,\mathrm{m}$, while decreasing AF by $2.02$ points. Within the diversity route, DBS-Hybrid exceeds DBS by $0.65$ points in FAA, $1.14$ points in BWT, $0.36$ points in C1, $0.91$ points in C3, and $0.82$ points in final all-mission Recall@$300\,\mathrm{m}$, while decreasing AF by $1.03$ points. DBS is higher by $0.01$ points in FWT and $0.39$ points in C2. These comparisons support combining diversity-based selection with the representative--trimming--coverage--backfill allocation used by DBS-Hybrid.

\subsubsection{Visualization of Retained-Sample Distribution}
\label{sec:tsne_vis}

\begin{figure}[!t]
	\centering
	\includegraphics[width=0.95\linewidth]{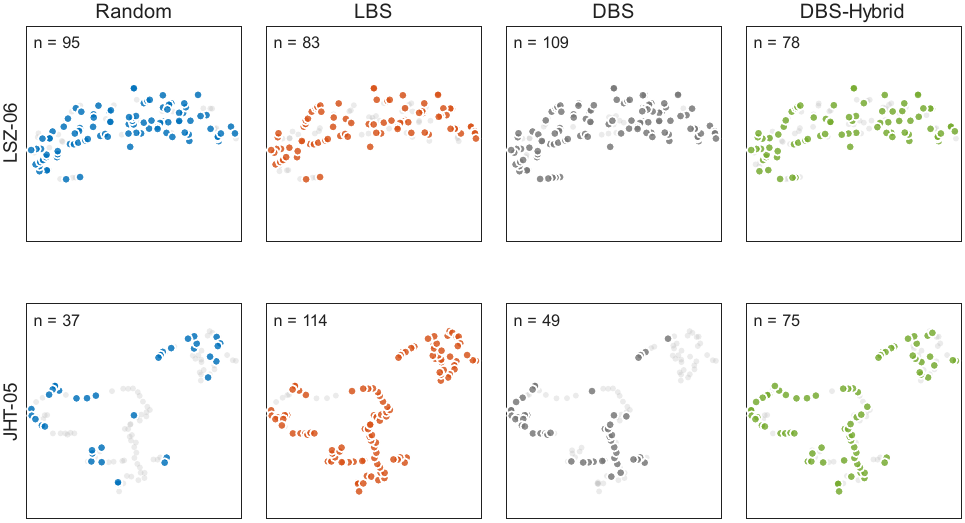}
	\caption{\textbf{t-SNE visualization of retained samples after the Forward mission order ($k=10$, seed 0).} Light-gray points denote all samples from each displayed mission, and highlighted points denote samples retained in the final ER buffer. Columns compare Random, LBS, DBS, and DBS-Hybrid for \texttt{LSZ-06} and \texttt{JHT-05} using the same fixed DINOv2+GeM representation. The annotation $n$ gives the number of retained samples from the corresponding mission. The visualization is qualitative; distances in the two-dimensional t-SNE embedding should not be interpreted as exact distances in the original feature space.}
	\label{fig:tsne_comparison}
\end{figure}

Figure~\ref{fig:tsne_comparison} visualizes the retained samples at the final sequential step ($k=10$) using t-SNE. It shows \texttt{LSZ-06}, an early mission retained throughout the sequence, and \texttt{JHT-05}, whose projected samples form several separated regions. The visualization uses seed 0.

To place all methods in a common representation space, feature vectors are extracted using the original pre-trained DINOv2 backbone~\cite{dinov22023} with the fixed GeM pooling aggregator. The same fixed feature extractor is used for all methods, so the displayed differences arise from the samples retained by each replay strategy. For each mission, the complete sample set and the corresponding retained subset are projected jointly into two dimensions using t-SNE.

The replay strategies produce different retained-sample distributions. For \texttt{LSZ-06}, all four strategies retain samples along most of the dominant projected structure. For \texttt{JHT-05}, Random and DBS retain comparatively sparse subsets, LBS retains the largest subset, and DBS-Hybrid distributes 75 retained samples across several separated regions. The contrast between LBS and DBS reflects the selection criterion under Min-Guar, whereas the contrast between DBS and DBS-Hybrid shows the retained-sample patterns produced by the two diversity-oriented selectors under Min-Guar and Hybrid allocation, respectively. These patterns complement the quantitative results in Table~\ref{tab:utility-operator-interaction}.

\subsubsection{Sensitivity to Buffer Capacity}
\label{sec:buffer_sensitivity}

\begin{figure}[!t]
	\centering
	\includegraphics[width=0.95\linewidth]{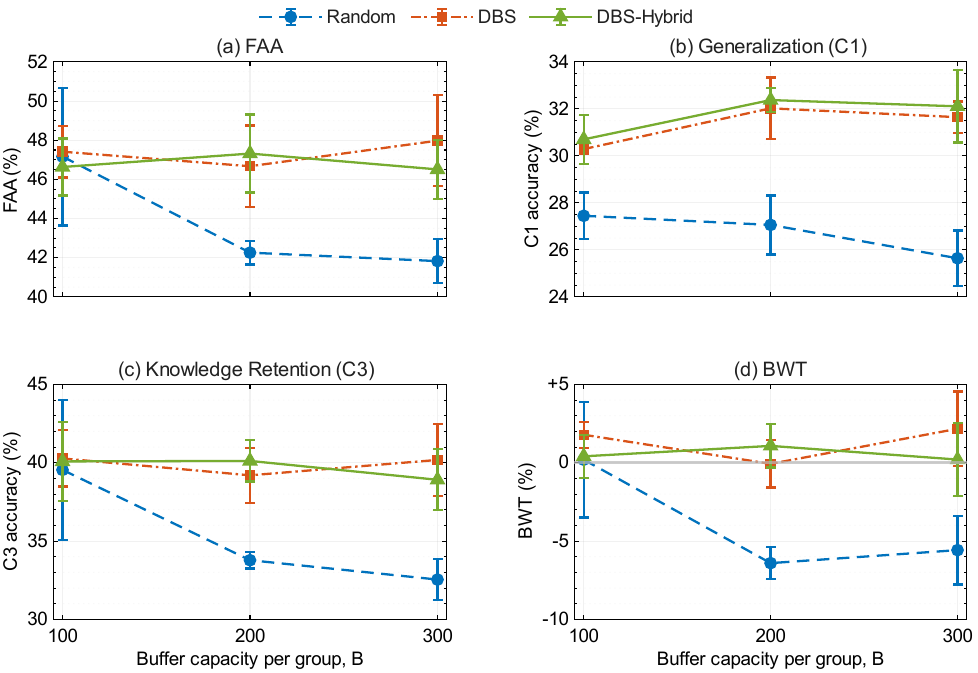}
	\caption{\textbf{Sensitivity to the per-group ER-buffer capacity $B$.} The four panels report FAA, generalization (C1), knowledge retention (C3), and BWT for $B\in\{100,200,300\}$. With $N=2$, the corresponding total dynamic replay upper bounds are $N^2B\in\{400,800,1200\}$ airborne images. Points show three-seed means and error bars show sample standard deviations.}
	\label{fig:sensitivity_combined}
\end{figure}

We examine the sensitivity of replay performance to the per-group ER-buffer capacity $B\in\{100,200,300\}$ using Random, DBS, and DBS-Hybrid. Since the D\&C grouping modulus is $N=2$, the four classifier groups give total dynamic replay upper bounds of $N^2B\in\{400,800,1200\}$ airborne images. The primary experiments use $B=200$, corresponding to a total dynamic replay upper bound of 800 images. We evaluate FAA, generalization (C1), knowledge retention (C3), and BWT; the results are summarized in Fig.~\ref{fig:sensitivity_combined}.

Across all three tested capacities, both DBS and DBS-Hybrid have higher mean C1, C3, and BWT than Random. Random decreases when the capacity increases from $B=100$ to $B=200$ and remains lower at $B=300$, whereas the two diversity-based selectors maintain more stable mean performance on these criteria. Replay composition therefore remains important as the available capacity changes.

DBS-Hybrid has higher mean C1 than DBS at all three tested capacities. At the default setting $B=200$, it also gives higher mean FAA, C3, and BWT. Thus, DBS-Hybrid consistently gives higher C1 than DBS across the tested capacities and also gives higher FAA, C3, and BWT at the primary replay budget.

\subsection{Computational and Storage Analysis}
\label{sec:efficiency}

The complete satellite reference dataset $\mathcal{D}_{sat}$ contains 2,037,473 tiles and occupies approximately 97.05\,GB. Before the mission sequence, it is used to train the initial model and construct the static satellite exemplar memory $\mathcal{M}_{EX}$. Continual adaptation uses $\mathcal{M}_{EX}$, which contains 91,747 unique satellite images occupying approximately 4.93\,GB, together with the trained model of approximately 372\,MB. Under the default capacity $B=200$, the dynamic ER buffer contains at most $N^2B=800$ airborne images and occupies approximately 19--21\,MB. At query time, the predicted geographic class is converted directly to its grid-cell centre. The satellite storage scales with the mapped area and geographic label space, while the dynamic replay storage is fixed by $B$ and $N$.

For a per-group candidate pool of size $n$ with $d$-dimensional feature representations, uniform random buffer update has $O(n)$ time complexity. Dense DBS scoring requires $O(n^2d)$ time and $O(n^2)$ pairwise-similarity storage in the worst case. DBS-Hybrid uses the same pairwise feature relationships and has the same worst-case complexities. We measure its selection time at 37 buffer updates, with three repetitions per update and 111 measurements in total. The mean, median, 95th-percentile, and maximum selection times are 65.893, 45.818, 224.058, and 399.053\,ms, respectively. These measurements isolate sample selection from feature preparation and buffer construction.

\begin{table*}[t]
	\centering
	\caption{Measured workstation resource costs under the Forward mission order. Continual-adaptation values are mean $\pm$ sample standard deviation over three seeds. Peak GPU allocated memory is the maximum allocation recorded during adaptation. The DBS-Hybrid replay-buffer update time includes candidate feature preparation, sample selection, and buffer update. Query-time measurements use the trained DBS-Hybrid model with batch size one; image loading and host-to-device transfer are included, while model loading is excluded.}
	\label{tab:resource-cost}
	\setlength{\tabcolsep}{4.5pt}
	\resizebox{\textwidth}{!}{%
		\begin{tabular}{lrrrr}
			\toprule
			\multicolumn{5}{l}{\textbf{(a) Ten-step continual adaptation}} \\
			Method & Total time (h) & Mean time/step (min) & Peak GPU allocated (MiB) & Replay-buffer update / adaptation time (\%) \\
			\midrule
			Random & 0.436$\pm$0.002 & 2.614$\pm$0.011 & 1418.68$\pm$0.00 & -- \\
			LBS & 0.414$\pm$0.010 & 2.486$\pm$0.061 & 1429.71$\pm$0.00 & -- \\
			DBS & 0.438$\pm$0.004 & 2.629$\pm$0.023 & 1429.71$\pm$0.00 & -- \\
			DBS-Hybrid & 0.444$\pm$0.004 & 2.665$\pm$0.022 & 1424.70$\pm$0.00 & 1.455$\pm$0.078 \\
			\midrule
			\multicolumn{5}{l}{\textbf{(b) DBS-Hybrid query-time inference}} \\
			Evaluation set & Mean latency/query (ms) & p50/p95 (ms) & Throughput (query/s) & Peak GPU allocated (MiB) \\
			\midrule
			Final all-mission geo-localization & 11.281$\pm$2.604 & 10.277/15.642 & 88.642 & 372.94 \\
			Held-out airborne missions (C1) & 11.306$\pm$2.387 & 10.447/15.528 & 88.447 & 372.94 \\
			\bottomrule
		\end{tabular}
	}
\end{table*}

Table~\ref{tab:resource-cost} summarizes the workstation measurements for Random, LBS, DBS, and DBS-Hybrid. Across the four replay strategies, the ten-step Forward adaptation requires $0.414$--$0.444$\,h, or approximately $2.49$--$2.67$\,min per sequential step, with peak allocated GPU memory between $1418.68$ and $1429.71$\,MiB. DBS-Hybrid requires $0.444\pm0.004$\,h for the complete sequence and $2.665\pm0.022$\,min per step. Its complete replay-buffer update accounts for $1.455\pm0.078\%$ of the total adaptation time and includes candidate feature preparation, sample selection, and buffer update.

At batch size one, DBS-Hybrid requires $11.281\pm2.604$\,ms per final all-mission query and $11.306\pm2.387$\,ms per held-out query, corresponding to throughputs of approximately 88.6 and 88.4 queries/s, respectively. The latency includes image loading and host-to-device transfer and excludes model loading. All measurements in Table~\ref{tab:resource-cost} use the RTX~4090 workstation described in Sec.~\ref{sec:impl_details}.

\subsection{Qualitative Geo-localization Analysis}
\label{sec:qualitative}

\begin{figure}[p]
	\centering
	{\setlength{\tabcolsep}{3pt}
		\begin{tabular}{@{}cccc@{}}
			{\footnotesize\bfseries Airborne query} & {\footnotesize\bfseries Ground-truth tile} & {\footnotesize\bfseries Random} & {\footnotesize\bfseries DBS-Hybrid} \\
			\multicolumn{4}{l}{{\footnotesize\bfseries (a) Final all-mission geo-localization, VIS}} \\
			\includegraphics[width=0.16\textwidth,height=0.16\textwidth]{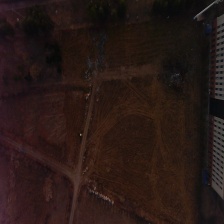} &
			\includegraphics[width=0.16\textwidth,height=0.16\textwidth]{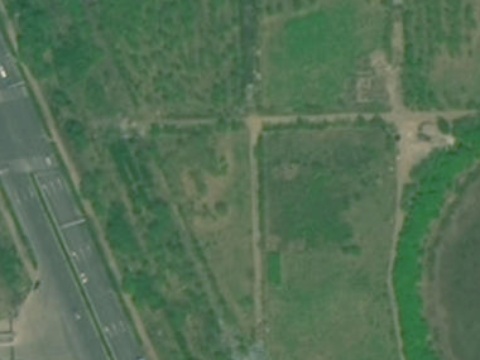} &
			\includegraphics[width=0.16\textwidth,height=0.16\textwidth]{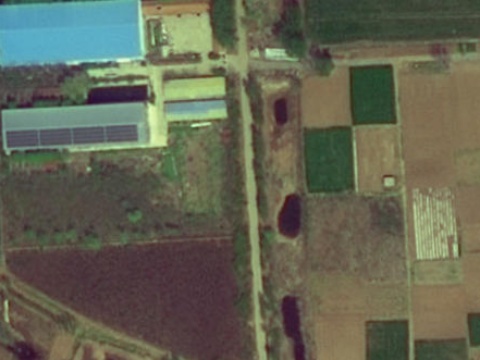} &
			\includegraphics[width=0.16\textwidth,height=0.16\textwidth]{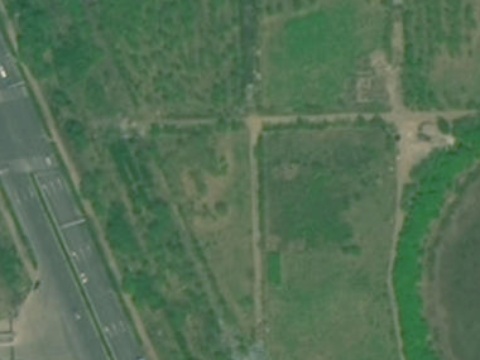} \\
			& & \scriptsize\shortstack{Median: 34523.3\,m\\Recall@300\,m: 0/3} & \scriptsize\shortstack{Median: 50.9\,m\\Recall@300\,m: 3/3} \\[1mm]
			\multicolumn{4}{l}{{\footnotesize\bfseries (b) Final all-mission geo-localization, IR}} \\
			\includegraphics[width=0.16\textwidth,height=0.16\textwidth]{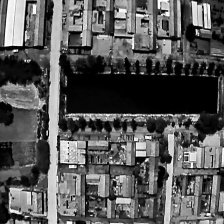} &
			\includegraphics[width=0.16\textwidth,height=0.16\textwidth]{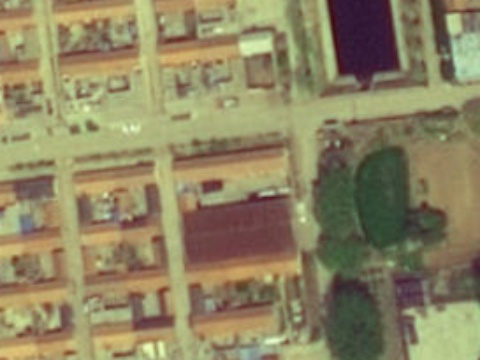} &
			\includegraphics[width=0.16\textwidth,height=0.16\textwidth]{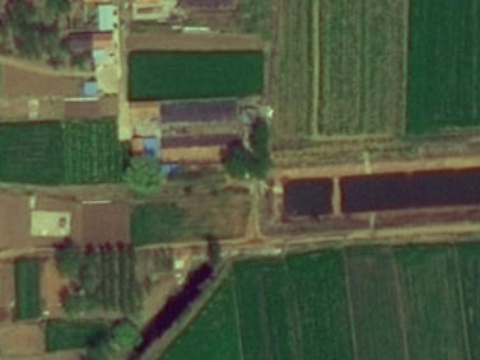} &
			\includegraphics[width=0.16\textwidth,height=0.16\textwidth]{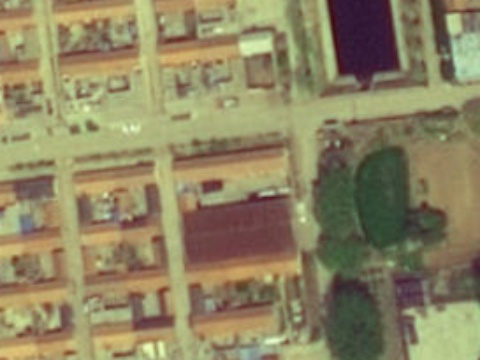} \\
			& & \scriptsize\shortstack{Median: 1437.7\,m\\Recall@300\,m: 0/3} & \scriptsize\shortstack{Median: 38.4\,m\\Recall@300\,m: 3/3} \\[1mm]
			\multicolumn{4}{l}{{\footnotesize\bfseries (c) Generalization (C1), VIS}} \\
			\includegraphics[width=0.16\textwidth,height=0.16\textwidth]{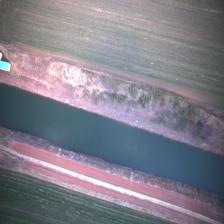} &
			\includegraphics[width=0.16\textwidth,height=0.16\textwidth]{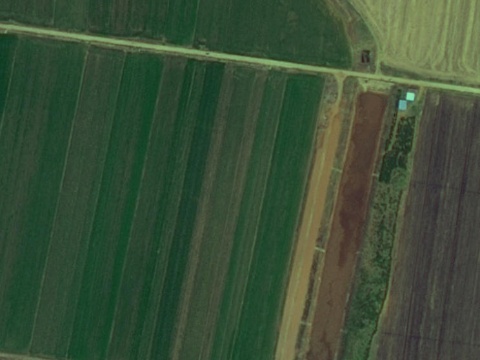} &
			\includegraphics[width=0.16\textwidth,height=0.16\textwidth]{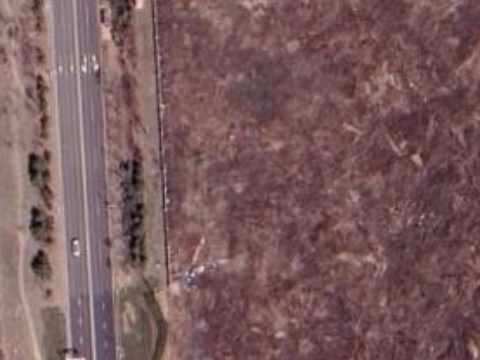} &
			\includegraphics[width=0.16\textwidth,height=0.16\textwidth]{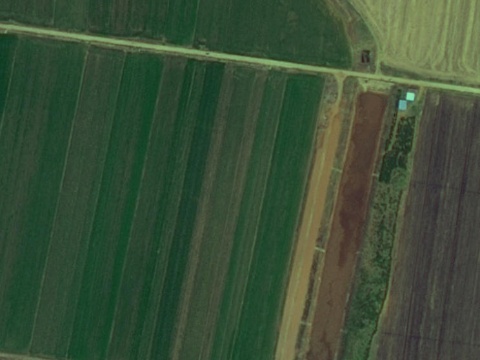} \\
			& & \scriptsize\shortstack{Median: 35270.3\,m\\Recall@300\,m: 0/3} & \scriptsize\shortstack{Median: 58.0\,m\\Recall@300\,m: 3/3} \\[1mm]
			\multicolumn{4}{l}{{\footnotesize\bfseries (d) Generalization (C1), IR}} \\
			\includegraphics[width=0.16\textwidth,height=0.16\textwidth]{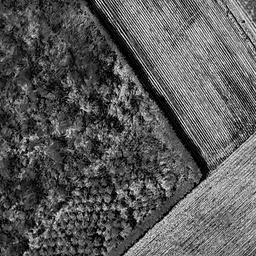} &
			\includegraphics[width=0.16\textwidth,height=0.16\textwidth]{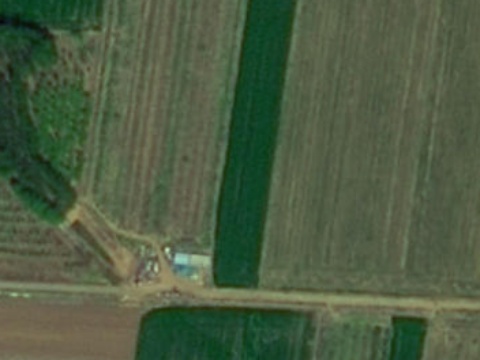} &
			\includegraphics[width=0.16\textwidth,height=0.16\textwidth]{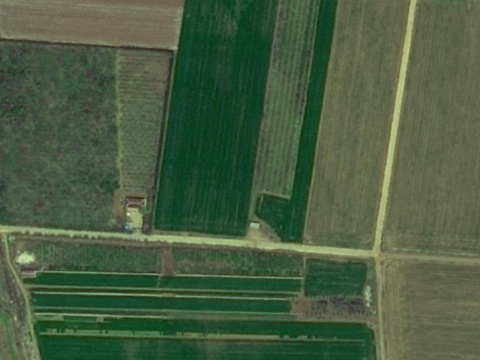} &
			\includegraphics[width=0.16\textwidth,height=0.16\textwidth]{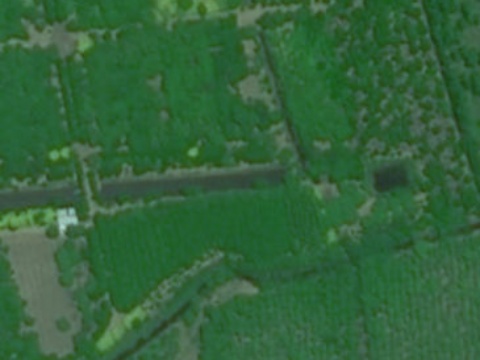} \\
			& & \scriptsize\shortstack{Median: 125.9\,m\\Recall@300\,m: 3/3} & \scriptsize\shortstack{Median: 24943.1\,m\\Recall@300\,m: 1/3}
	\end{tabular}}
	\caption{\textbf{Qualitative geo-localization cases under the Forward mission order.} Each prediction column displays the satellite tile produced by the seed with the median localization error among three seeds; the annotations report this median error and the number of Recall@$300\,\mathrm{m}$ hits across the three seeds. Panels (a)--(c) show cases in which DBS-Hybrid is within $300\,\mathrm{m}$ for all three seeds while Random is outside $300\,\mathrm{m}$ for all three. Panel (d) provides the complementary failure case, in which Random succeeds for all three seeds while DBS-Hybrid succeeds for one of the three seeds.}
	\label{fig:qualitative_cases}
\end{figure}

Figure~\ref{fig:qualitative_cases} presents four query-level cases from the recorded Forward predictions: three cases in which DBS-Hybrid succeeds across all seeds and one complementary case in which Random succeeds more consistently. The cases include VIS and IR airborne observations from final all-mission geo-localization and generalization (C1). Each row shows the airborne query, its ground-truth VIS satellite tile, and the predicted satellite tiles from Random and DBS-Hybrid. For each method, the displayed prediction corresponds to the seed with the median localization error, and the annotation gives this error and the number of predictions within $300\,\mathrm{m}$. The IR rows evaluate IR airborne queries against VIS satellite reference images.

Panels (a)--(c) in Fig.~\ref{fig:qualitative_cases} show geographically consistent DBS-Hybrid predictions across all three seeds under different sensor modalities and airborne--satellite appearances. Together with Panel (d), these cases link the multi-threshold recall and position-error statistics in Table~\ref{tab:table4-multithreshold-forward-3seed} to query-level outcomes.

\begin{figure}[t]
	\centering
	{\setlength{\tabcolsep}{3pt}
		\begin{tabular}{@{}cccc@{}}
			{\footnotesize\bfseries Airborne query} & {\footnotesize\bfseries Ground-truth tile} & {\footnotesize\bfseries Random} & {\footnotesize\bfseries DBS-Hybrid} \\
			\includegraphics[width=0.18\textwidth,height=0.18\textwidth]{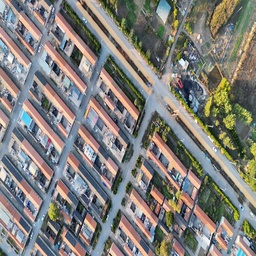} &
			\includegraphics[width=0.18\textwidth,height=0.18\textwidth]{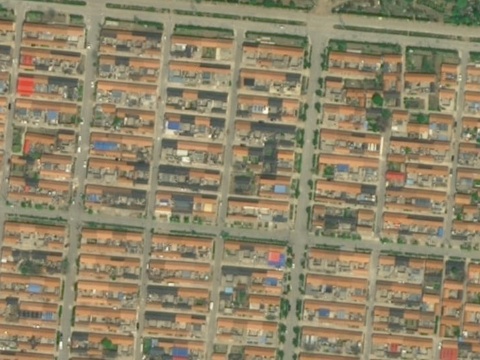} &
			\includegraphics[width=0.18\textwidth,height=0.18\textwidth]{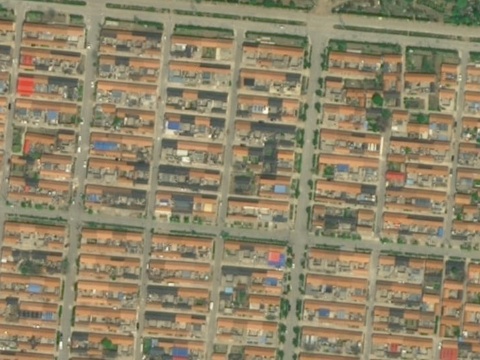} &
			\includegraphics[width=0.18\textwidth,height=0.18\textwidth]{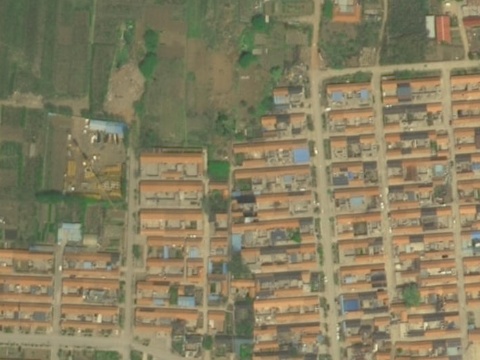} \\
			& & \scriptsize\shortstack{Median: 72.1\,m\\Recall@300\,m: 3/3} & \scriptsize\shortstack{Median: 3454.0\,m\\Recall@300\,m: 0/3}
	\end{tabular}}
	\caption{\textbf{Geo-localization failure case under repetitive aerial appearance.} The generalization (C1) VIS query, its ground-truth satellite region, and the geographically distant DBS-Hybrid prediction contain visually similar repeated rows of low-rise buildings. Random remains within $300\,\mathrm{m}$ for all three seeds, whereas DBS-Hybrid is outside $300\,\mathrm{m}$ for all three seeds, illustrating residual perceptual aliasing between geographically distinct regions.}
	\label{fig:repetitive_case}
\end{figure}

Figure~\ref{fig:repetitive_case} isolates perceptual aliasing caused by repetitive aerial appearance. The airborne query and ground-truth satellite region contain repeated rows of dense low-rise buildings, and the geographically distant DBS-Hybrid prediction contains a similar structural pattern. The example shows how visually similar structures can connect distinct geographic grid cells.

\FloatBarrier

\section{Discussion}
\label{sec:discussion}

\subsection{Heterogeneous Memory and Replay Selection}
\label{sec:discussion_replay}

The heterogeneous memory framework assigns fixed satellite information and changing mission experience to different memory components. Before the mission sequence, the satellite reference dataset trains the initial model and provides the class-balanced static satellite exemplar memory. During continual adaptation, the bounded replay buffer retains selected airborne observations from previous missions. This separation preserves fixed geographic supervision while limiting the stored airborne experience.

Replay selection determines which airborne observations remain available for later missions. Herding emphasizes class central tendency, LBS emphasizes prediction difficulty, and DBS emphasizes low feature redundancy. A geographic class can contain complementary appearances across viewpoint, illumination, season, modality, and structural change. DBS-Hybrid combines prototype-based trimming with prototype-nearest representatives and cosine $k$-center coverage to retain such variation under the replay budget. Tables~\ref{tab:table6-allocation}--\ref{tab:utility-operator-interaction} support the combination of diversity-oriented selection and Hybrid allocation used by DBS-Hybrid, while Fig.~\ref{fig:tsne_comparison} shows the corresponding differences in the retained subsets.

\subsection{Performance across Metrics, Mission Orders, and Sensor Modalities}
\label{sec:discussion_tradeoffs}

Under the Forward mission order, DBS-Hybrid gives the highest FAA, C1, C3, and multi-threshold geo-localization recall among the evaluated methods. Its joint C1 and C3 results show that diversity-aware replay supports held-out airborne appearances while retaining knowledge from historical missions. C2, BWT, FWT, and AF provide complementary views of current-mission adaptation and performance changes on historical and future tasks.

Mission order changes the sequence of visual distribution shifts and the duration for which retained samples support subsequent adaptation. Across the five additional random orders, DBS-Hybrid ranks second in mean FAA, BWT, C1, and C3 and gives the highest mean FWT. The modality results further show the highest final VIS recall and the highest C1 recall for both VIS and IR queries. Because the Forward order ends with the 500-image IR mission \texttt{TD-13-seq}, the final IR result reflects both sensor modality and the final training mission. Overall, DBS-Hybrid gives the highest FAA, C1, C3, and multi-threshold recall under the primary protocol and remains among the leading methods across the additional mission orders.

\subsection{Limitations and Future Work}
\label{sec:limitations}

The benchmark contains 21 missions from two districts of Qingdao. Extending the evaluation to additional mapped regions, seasons, and sensor modalities will provide a broader assessment of mission-based DIL. The framework also relies on a satellite reference dataset acquired before deployment. Substantial changes in buildings, roads, or land cover can weaken its correspondence with current airborne observations, motivating selective map updates during long-term operation.

The current missions mainly contain nadir and low-oblique views, with off-nadir angles up to approximately $30^\circ$. Larger viewpoint changes introduce stronger projective distortions between airborne observations and satellite images. Geometric alignment or perspective-aware representations could extend the framework to these conditions. Repetitive aerial structures remain a source of geographically distant errors. Combining replay selection with local geometric verification or temporal consistency could help distinguish visually similar grid cells.

Prolonged hovering or dense frame extraction can produce near-identical observations before replay selection. Motion-aware filtering based on flight-control or visual-inertial information could remove redundant frames before continual adaptation. The present resource analysis is based on workstation measurements. Future work will evaluate adaptation time, memory use, and energy consumption on embedded UAV hardware.

\section{Conclusion}
\label{sec:conclusion}

This work formulates aerial VPR in mapped operating areas as mission-based domain-incremental learning, where the geographic label space remains fixed while airborne observations change across successive missions. We develop a heterogeneous memory framework that separates the static satellite exemplar memory from bounded replay of airborne experience. Within the replay buffer, LBS and DBS define utility by prediction difficulty and feature-space diversity, respectively; DBS-Hybrid follows the diversity route with prototype-based trimming and representative-first feature-space coverage. Under the primary Forward order, DBS-Hybrid achieves the highest FAA, Generalization (C1), Knowledge Retention (C3), and multi-threshold geo-localization recall among the evaluated methods. The criterion, allocation, and component experiments support the DBS-Hybrid design. Together, heterogeneous memory and DBS-Hybrid provide a practical approach to continual aerial VPR under bounded airborne replay.

\section*{Data and code availability}
Dataset documentation and the satellite-map archive are currently available to the editor and reviewers through a private Zenodo review record (record 19394067). Upon acceptance, the authors will publicly release the benchmark documentation, split manifests, evaluation code, and author-owned data that can legally be redistributed. The current implementation is available at \url{https://github.com/VictoireWood/Aerial-Continual-Learning}. Third-party and partner-contributed assets remain subject to their original licences and agreements.

\appendix

\section{Exact Mission Sequences}
\label{sec:appendix_curriculum}

As discussed in Sec.~\ref{sec:curriculum}, the results vary with mission order. Tables~\ref{tab:curriculum_orders} and~\ref{tab:random_curriculum_orders} therefore list the exact mission progression for the four named sequences and five fixed random permutations.

\begin{table*}[htbp]
	\centering
	\renewcommand{\arraystretch}{1.2}
	\setlength{\tabcolsep}{12pt}
	\caption{\textbf{Exact mission sequences for the four named mission orders.} The indices $k=1 \dots 10$ identify the sequential step of each mission in the continual mission sequence.}
	\label{tab:curriculum_orders}
	\resizebox{\linewidth}{!}{
	\begin{tabular}{c l l l l}
		\toprule
		\textbf{Step $k$} & \textbf{Forward} & \textbf{Backward} & \textbf{Pressure} & \textbf{Robust} \\
		\midrule
		1 & \texttt{JHT-02} (VIS) & \texttt{TD-13-seq} (IR) & \texttt{TD-13-seq} (IR) & \texttt{JHT-02} (VIS) \\
		2 & \texttt{LSZ-07} (VIS) & \texttt{JHT-05} (IR) & \texttt{JHT-02} (VIS) & \texttt{TD-02-seq} (VIS) \\
		3 & \texttt{LSZ-06} (VIS) & \texttt{JHT-01} (VIS) & \texttt{JHT-05} (IR) & \texttt{TD-13-seq} (IR) \\
		4 & \texttt{TD-02-seq} (VIS) & \texttt{LSZ-01} (VIS) & \texttt{LSZ-07} (VIS) & \texttt{LSZ-07} (VIS) \\
		5 & \texttt{TD-07-seq} (VIS) & \texttt{JHT-04} (VIS) & \texttt{TD-02-seq} (VIS) & \texttt{JHT-04} (VIS) \\
		6 & \texttt{JHT-04} (VIS) & \texttt{TD-07-seq} (VIS) & \texttt{LSZ-06} (VIS) & \texttt{JHT-05} (IR) \\
		7 & \texttt{LSZ-01} (VIS) & \texttt{TD-02-seq} (VIS) & \texttt{JHT-04} (VIS) & \texttt{LSZ-06} (VIS) \\
		8 & \texttt{JHT-01} (VIS) & \texttt{LSZ-06} (VIS) & \texttt{TD-07-seq} (VIS) & \texttt{TD-07-seq} (VIS) \\
		9 & \texttt{JHT-05} (IR) & \texttt{LSZ-07} (VIS) & \texttt{LSZ-01} (VIS) & \texttt{LSZ-01} (VIS) \\
		10 & \texttt{TD-13-seq} (IR) & \texttt{JHT-02} (VIS) & \texttt{JHT-01} (VIS) & \texttt{JHT-01} (VIS) \\
		\bottomrule
	\end{tabular}
}
\end{table*}

\begin{table*}[htbp]
	\centering
	\renewcommand{\arraystretch}{1.15}
	\setlength{\tabcolsep}{3.5pt}
	\caption{\textbf{Exact mission sequences for the five fixed random-order evaluations.} Each row gives the mission at sequential step $k$ for one random mission order.}
	\label{tab:random_curriculum_orders}
	\resizebox{\linewidth}{!}{
	\begin{tabular}{c l l l l l}
		\toprule
		\textbf{Step $k$} & \textbf{ord-104729} & \textbf{ord-130363} & \textbf{ord-155921} & \textbf{ord-181081} & \textbf{ord-208367} \\
		\midrule
		1  & \texttt{JHT-04} & \texttt{TD-02-seq} & \texttt{TD-13-seq} & \texttt{TD-13-seq} & \texttt{JHT-02} \\
		2  & \texttt{JHT-02} & \texttt{LSZ-01} & \texttt{LSZ-07} & \texttt{TD-02-seq} & \texttt{TD-13-seq} \\
		3  & \texttt{LSZ-01} & \texttt{JHT-01} & \texttt{LSZ-06} & \texttt{LSZ-01} & \texttt{TD-02-seq} \\
		4  & \texttt{JHT-05} & \texttt{TD-07-seq} & \texttt{JHT-04} & \texttt{JHT-01} & \texttt{LSZ-01} \\
		5  & \texttt{LSZ-07} & \texttt{LSZ-07} & \texttt{JHT-02} & \texttt{JHT-05} & \texttt{TD-07-seq} \\
		6  & \texttt{TD-13-seq} & \texttt{JHT-02} & \texttt{TD-02-seq} & \texttt{JHT-04} & \texttt{JHT-01} \\
		7  & \texttt{LSZ-06} & \texttt{JHT-04} & \texttt{JHT-05} & \texttt{JHT-02} & \texttt{JHT-05} \\
		8  & \texttt{TD-07-seq} & \texttt{JHT-05} & \texttt{TD-07-seq} & \texttt{LSZ-06} & \texttt{LSZ-07} \\
		9  & \texttt{JHT-01} & \texttt{LSZ-06} & \texttt{LSZ-01} & \texttt{LSZ-07} & \texttt{LSZ-06} \\
		10 & \texttt{TD-02-seq} & \texttt{TD-13-seq} & \texttt{JHT-01} & \texttt{TD-07-seq} & \texttt{JHT-04} \\
		\bottomrule
	\end{tabular}}
\end{table*}

\section{Detailed Results for Additional Random Mission Orders}
\label{sec:appendix_random_orders}

The five additional random mission orders used in Sec.~\ref{sec:robustness} were fixed before evaluation. Each order was evaluated once for each of the six methods included in the mission-order analysis. Therefore, the individual-order entries below are single-order results, while the mean and sample standard deviation in the final column summarize variation across the five mission orders rather than repeated-seed variability. Table~\ref{tab:appendix-random-orders-faa-bwt-fwt-af} reports FAA and BWT, and Table~\ref{tab:appendix-random-orders-c1-c2-c3} reports C1--C3.

\begin{table*}[t]
	\centering
	\caption{Detailed FAA / BWT / FWT / AF (\%) results across five additional fixed random mission orders. Each order is evaluated once per method, and each cell reports FAA / BWT / FWT / AF, with lower values preferred only for AF. AF is averaged over the first $K-1$ tasks. The final column reports the mean and sample standard deviation across the five orders. Within each order and metric, best, second-best, and third-best values are shown in bold, underlined, and italic type, respectively.}
	\label{tab:appendix-random-orders-faa-bwt-fwt-af}
	\resizebox{\textwidth}{!}{%
		\begin{tabular}{lcccccc}
			\toprule
			Method & ord-104729 & ord-130363 & ord-155921 & ord-181081 & ord-208367 & Across-order mean \\
			\midrule
			DIL-DER++
			& 48.22/\textit{1.08}/\textbf{4.63}/\textit{4.49}
			& \textbf{52.11}/\textbf{4.26}/\underline{1.13}/\textbf{2.32}
			& \textbf{51.03}/2.61/\textbf{3.95}/3.87
			& 48.26/-0.53/-1.81/4.90
			& \underline{52.34}/\textbf{6.39}/0.30/\textbf{0.66}
			& \textbf{50.39$\pm$2.03} / \textbf{2.76$\pm$2.70} / \textit{1.64$\pm$2.66} / \textbf{3.25$\pm$1.75} \\
			
			DIL-iCaRL
			& 42.20/-6.98/\underline{3.85}/9.81
			& \underline{47.93}/\textit{-2.79}/-0.30/6.72
			& 47.46/-4.78/1.53/6.85
			& \textbf{51.95}/\textbf{2.64}/\textit{1.74}/\textbf{1.48}
			& 46.52/-2.83/-5.05/5.04
			& 47.21$\pm$3.49 / -2.95$\pm$3.57 / 0.35$\pm$3.36 / 5.98$\pm$3.04 \\
			
			Random
			& \underline{50.09}/0.68/\textit{2.88}/5.02
			& \textit{45.13}/-4.42/\textbf{3.16}/8.13
			& \underline{50.88}/\underline{4.05}/-4.92/\textbf{1.40}
			& 46.55/-2.96/\textbf{3.83}/5.70
			& 48.44/\textit{-0.81}/-3.24/\textit{2.91}
			& 48.22$\pm$2.40 / -0.69$\pm$3.29 / 0.34$\pm$4.10 / 4.63$\pm$2.60 \\
			
			LBS
			& 47.23/-2.22/0.87/6.40
			& 44.78/-3.42/-1.29/6.79
			& 46.81/-2.79/2.41/7.20
			& 47.35/-1.20/-0.02/\textit{3.67}
			& \textbf{53.40}/\underline{0.36}/\textit{0.81}/\underline{2.86}
			& 47.91$\pm$3.24 / -1.85$\pm$1.48 / 0.56$\pm$1.35 / 5.39$\pm$1.98 \\
			
			DBS
			& \textit{48.85}/\underline{1.23}/1.40/\underline{2.76}
			& 42.88/-2.92/-0.36/\textit{6.53}
			& \textit{50.72}/\textbf{4.76}/\textit{2.81}/\underline{1.54}
			& \underline{49.00}/\textit{0.37}/\underline{1.84}/\underline{2.85}
			& \textit{50.43}/-5.79/\textbf{3.01}/7.92
			& \textit{48.38$\pm$3.19} / \textit{-0.47$\pm$4.04} / \underline{1.74$\pm$1.35} / \textit{4.32$\pm$2.75} \\
			
			DBS-Hybrid
			& \textbf{50.32}/\textbf{3.35}/2.06/\textbf{2.47}
			& 45.05/\underline{0.38}/\textit{0.77}/\underline{4.13}
			& 50.17/\textit{2.83}/\underline{3.54}/\textit{2.47}
			& \textit{48.34}/\underline{0.57}/0.85/3.68
			& 49.41/-4.05/\underline{2.26}/6.07
			& \underline{48.66$\pm$2.16} / \underline{0.61$\pm$2.92} / \textbf{1.90$\pm$1.14} / \underline{3.76$\pm$1.48} \\
			\bottomrule
		\end{tabular}%
	}
\end{table*}

\begin{table*}[t]
	\centering
	\caption{Detailed C1 / C2 / C3 (\%) results across five additional fixed random mission orders. Each order is evaluated once per method. The final column reports the mean and sample standard deviation across the five orders. Within each order and criterion, best, second-best, and third-best values are shown in bold, underlined, and italic type, respectively.}
	\label{tab:appendix-random-orders-c1-c2-c3}
	\resizebox{\textwidth}{!}{%
		\begin{tabular}{lcccccc}
			\toprule
			Method & ord-104729 & ord-130363 & ord-155921 & ord-181081 & ord-208367 & Across-order mean \\
			\midrule
			DIL-DER++ & \textbf{34.97}/47.25/48.18 & \underline{30.57}/\textit{48.27}/\textbf{42.49} & \underline{32.68}/\textit{48.68}/\textbf{51.69} & \textit{32.24}/\textit{48.74}/47.56 & \textit{30.64}/46.59/\underline{57.03} & \textbf{32.22$\pm$1.80} / 47.91$\pm$0.95 / \textbf{49.39$\pm$5.39} \\
			DIL-iCaRL & 30.42/\textit{48.48}/44.46 & \textbf{31.62}/\textbf{50.45}/\underline{39.92} & \textit{32.39}/\textbf{51.76}/47.60 & \textit{31.62}/\textbf{49.57}/\textbf{50.29} & \underline{30.90}/49.07/48.59 & \textit{31.39$\pm$0.76} / \textbf{49.87$\pm$1.28} / 46.17$\pm$4.09 \\
			Random & 29.33/\textbf{49.48}/\textbf{54.69} & 28.28/\underline{49.10}/36.22 & \textbf{32.71}/47.24/\underline{51.53} & 25.40/\underline{49.21}/46.23 & \textbf{31.33}/49.17/49.53 & 29.41$\pm$2.83 / \textit{48.84$\pm$0.91} / 47.64$\pm$7.09 \\
			LBS & 31.37/49.22/48.01 & 28.53/47.85/36.22 & 29.69/49.32/46.81 & 28.13/48.43/46.31 & 27.98/\underline{53.08}/\textbf{57.29} & 29.14$\pm$1.41 / \underline{49.58$\pm$2.04} / 46.93$\pm$7.48 \\
			DBS & \textit{33.19}/47.75/\textit{52.66} & 28.60/45.50/34.51 & 30.82/46.44/\textit{51.06} & \underline{33.37}/48.67/\underline{48.80} & 30.17/\textbf{55.64}/\textit{54.56} & 31.23$\pm$2.04 / 48.80$\pm$4.01 / \textit{48.32$\pm$8.01} \\
			DBS-Hybrid & \underline{34.39}/47.31/\underline{53.42} & \textit{29.77}/44.71/\textit{36.88} & 31.44/47.63/50.75 & \textbf{34.39}/47.83/\textit{48.30} & 30.39/\textit{53.06}/53.96 & \underline{32.07$\pm$2.20} / 48.11$\pm$3.04 / \underline{48.66$\pm$6.96} \\
			\bottomrule
		\end{tabular}%
	}
\end{table*}

\bibliographystyle{elsarticle-harv} 
\bibliography{harvref.bib}




\end{document}